\documentclass{article}

\PassOptionsToPackage{numbers, compress}{natbib}
\usepackage[main, final]{neurips_2026}

\usepackage[utf8]{inputenc} 
\usepackage[T1]{fontenc}    
\usepackage{hyperref}       
\usepackage{url}            
\usepackage{booktabs}       
\usepackage{amsfonts}       
\usepackage{nicefrac}       
\usepackage{microtype}      
\usepackage{xcolor}         
\usepackage{enumitem}
\usepackage{graphicx}      
\usepackage{multirow}      
\usepackage{threeparttable}
\usepackage[table]{xcolor}
\newcommand{\pos}[1]{\textcolor{green!45!black}{+#1}}
\newcommand{\negv}[1]{\textcolor{red!65!black}{-#1}}

\usepackage{booktabs}
\usepackage{multirow}
\usepackage{threeparttable}

\usepackage{amssymb}

\usepackage{wrapfig}

\usepackage{graphicx}
\usepackage{caption}
\usepackage{amsmath}

\usepackage{listings}
\usepackage{xcolor}
\lstdefinestyle{jsonstyle}{
    basicstyle=\ttfamily\scriptsize,
    breaklines=true,
    columns=fullflexible,
    frame=single,
    framerule=0.3pt,
    rulecolor=\color{gray!40},
    backgroundcolor=\color{gray!5},
    showstringspaces=false,
    keepspaces=true,
    tabsize=2
}

\usepackage[utf8]{inputenc}
\usepackage[T1]{fontenc}
\usepackage{hyperref}
\usepackage{url}
\usepackage{graphicx}
\usepackage{xcolor}
\usepackage{booktabs}
\usepackage{amsmath}
\usepackage{amsfonts}
\usepackage{microtype}

\usepackage[utf8]{inputenc}
\usepackage[T1]{fontenc}
\usepackage{hyperref}

\title{Revisiting Embodied Chain-of-Thought for Generalizable Robot Manipulation}

\author{%
\parbox{0.98\textwidth}{\centering\normalfont
\textbf{Nan Sun}$^{1,2}$ \quad
\textbf{Yuan Zhang}$^{2,\dagger,\star}$ \quad
\textbf{Yongkun Yang}$^{3}$ \quad
\textbf{Wentao Zhao}$^{1}$ \quad
\textbf{Peiyan Li}$^{2,4}$ \\
\textbf{Jun Guo}$^{1,2}$ \quad
\textbf{Wenxuan Song}$^{2,5}$ \quad
\textbf{Pengxiang Ding}$^{2,6}$ \quad
\textbf{Runze Suo}$^{2,7,9}$ \\
\textbf{Yifei Su}$^{2}$ \quad
\textbf{Xin Xiao}$^{8}$ \quad
\textbf{Xinghang Li}$^{2}$ \quad
\textbf{Huaping Liu}$^{1,\star}$ \\[2.5mm]
{\small
$^{1}$Tsinghua University \quad
$^{2}$Xiaomi Robotics \quad
$^{3}$Peking University \quad
$^{4}$CASIA \\
$^{5}$HKUST(GZ) \quad
$^{6}$Zhejiang University \quad
$^{7}$Fudan University \quad
$^{8}$Wuhan University \\
$^{9}$Shanghai Innovation Institute \quad
$^{\dagger}$Project Lead \quad
$^{\star}$Corresponding Author
}
}
}

\begin{document}

\maketitle

\begin{abstract}

Embodied chain-of-thought (CoT) aims to bridge linguistic reasoning with robotic control, yet its effective form and integration remain underexplored.
In this paper, we revisit embodied CoT for robotic control at an unprecedented scale.
We curate the largest embodied CoT corpus to date, comprising 978,743 trajectories, 226.3M samples, and 2592.5 hours of data.
Through extensive experiments, we show that effective CoT must ground high-level semantic understaning in concrete linguistic action guidance -- such as end-effector movement and image-space trajectories -- whereas high-level reasoning alone yields marginal gains.
More importantly, we identify that explicit CoT does not scale reliably as an autoregressive action prefix, suffering from compounding errors during inference.
To address these challenges, we propose ERVLA, a vision-language-action (VLA) model that effectively leverages linguistic reasoning in generalizable robot manipulation.
ERVLA is trained using a CoT-dropout strategy, allowing the model to leverage rich reasoning traces during training while predicting actions directly without CoT during inference to bypass autoregressive instability.
This approach enables reliable scaling with increasing pre-training data.
ERVLA achieves state-of-the-art results on LIBERO-Plus with an 86.9\% success rate and reaches 53.2\% on VLABench, showcasing superior performance in out-of-distribution settings.
Furthermore, ERVLA outperforms competitive state-of-the-art baselines in real-robot experiments, especially in handling semantic ambiguity and long-horizon tasks.
Code, data, and model checkpoints are available on our
\href{https://taoshuaiz.github.io/ERVLA/}{project website}.

\end{abstract}

\begin{figure}[t]
    \centering
    \includegraphics[width=\textwidth]{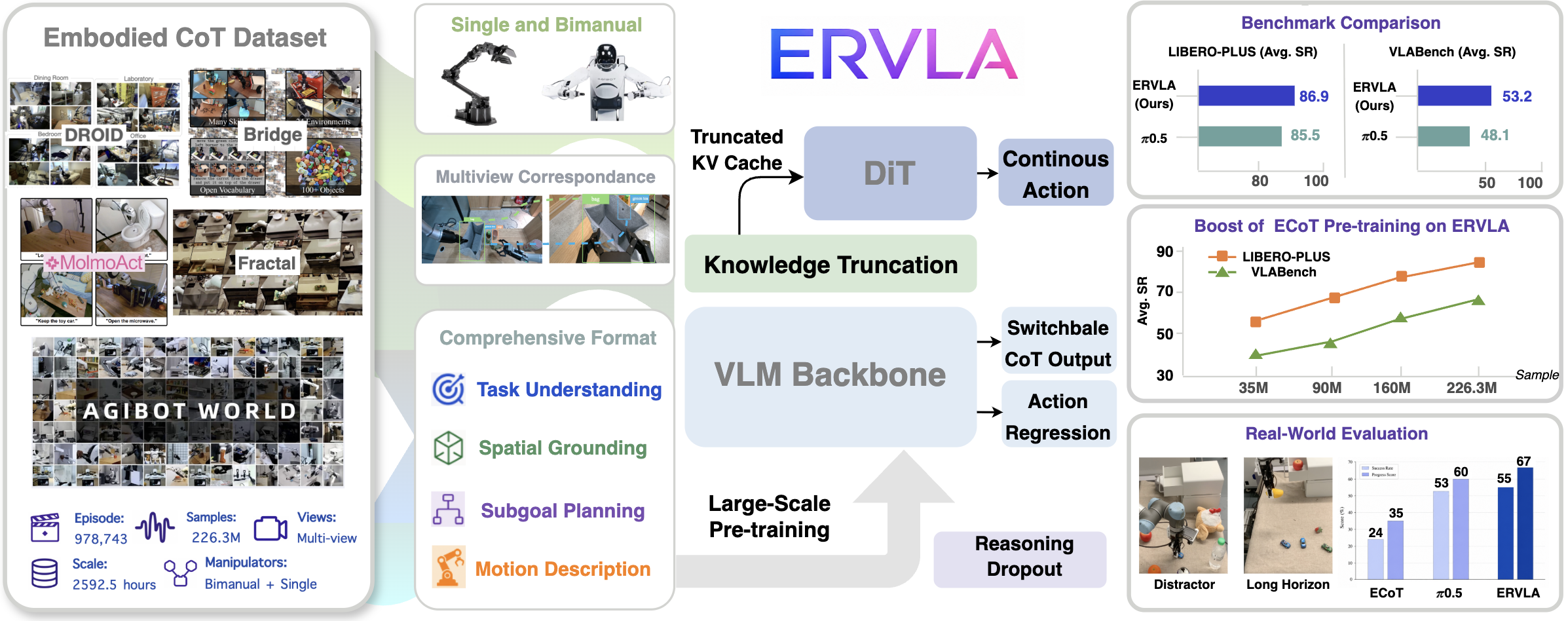}
\caption{
\textbf{Overview of ERVLA.}
We build the largest and most comprehensive embodied CoT corpus to date, which serves as the reasoning pre-training foundation for ERVLA. ERVLA uses CoT supervision, auxiliary action queries, knowledge-truncated KV conditioning, and reasoning dropout to internalize reasoning into action-aware VLM representations and predict continuous actions without mandatory test-time CoT. It achieves state-of-the-art performance on two simulation benchmarks, benefits from more embodied CoT data, and shows stronger real-world robustness under semantic ambiguity and long-horizon tasks.
}
    \label{fig:overview}

\vspace{-5mm}
    
\end{figure}

\section{Introduction}
\label{intro}

Vision-language-action (VLA) models have emerged as a promising paradigm for generalizable robot control by transferring large-scale visual and semantic priors from pre-trained vision-language models (VLMs) into embodied decision making~\cite{black2026pi0visionlanguageactionflowmodel,intelligence2025pi05visionlanguageactionmodelopenworld,kim2024openvla,zitkovich2023rt}.
By grounding language instructions in visual observations and mapping them to robot actions, these models provide a unified interface for open-vocabulary and long-horizon manipulation.
However, enhanced perception and broader semantic coverage do not inherently guarantee better action generation~\cite{zhang2026vlm4vla}.
In practice, VLA models often struggles in complex, out-of-distributed scenarios which demands deliberation prior to execution.
This bottleneck has motivated the development of reason-before-act policies, which leverages intermediate reasoning to guide action generation.
Previous embodied chain-of-thought (ECoT) methods~\cite{zawalski2024robotic} showcase that supervising explicit reasoning traces via next-token prediction can improve robustness and generalization.
Since then, the design space has rapidly expanded, encompassing diverse reasoning modalities, architectural designs for reasoning-action coupling, and specialized training recipes~\cite{chen2025trainingstrategiesefficientembodied}.

Despite the progress, \textbf{a systematic understanding of reasoning in VLA remains elusive.} 
The essence of ECoT is not to make robots ``say more,'' but to translate the semantic abstractions learned by VLMs from internet-scale data into intermediate representations that are genuinely useful for action generation.
Existing methods instantiate reasoning through various lenses, including scene understanding, subtask decomposition, spatial grounding, end-effector trajectory prediction, and future-frame prediction~\cite{kim2026cosmos,zawalski2024robotic,sun2024emmaxembodiedmultimodalaction,zhao2025cotvlavisualchainofthoughtreasoning,intelligence2025pi05visionlanguageactionmodelopenworld}.
However, these choices are tightly coupled with specific architectures and training objectives, making it difficult to identify what actually improves control. 
Consequently, the field lacks a principled understanding of which forms of embodied reasoning truly facilitate effective action generation.

\textbf{A second unresolved question is how should reasoning interact with the policy.} 
Early ECoT-style formulations~\cite{zawalski2024robotic,sun2024emmaxembodiedmultimodalaction} model reasoning as a prefix of action generation: the model first predicts embodied reasoning traces and then generates discretized action tokens through next-token prediction. 
This design is simple but slow and brittle, since long reasoning traces increase latency and action generation depends heavily on preceding reasoning, leading to compounding errors during inference.
Recent works decouple reasoning from action by distilling reasoning into latent plans~\cite{liu2026last0latentspatiotemporalchainofthought,huang2026fastthinkactefficientvisionlanguageactionreasoning}, conditioning diffusion-based action heads on selected reasoning states~\cite{huang2025thinkactvisionlanguageactionreasoningreinforced,tang2025mindhandpurposefulrobotic}, or learning from reasoning data during training while predicting actions without CoT at inference time~\cite{chen2025trainingstrategiesefficientembodied}. 
This raises another central question: should reasoning be taken as a visible trace, a latent variable, or a training signal that reshapes the policy representation?

\textbf{A third but equally important question is whether embodied CoT can effectively scale.} Recent VLA pre-training efforts have begun to include richer semantic or reasoning-related supervision~\cite{intelligence2025pi05visionlanguageactionmodelopenworld, tang2025mindhandpurposefulrobotic}.
However, the scaling behavior of CoT data remains unclear. 
Public reasoning-enhanced robotic datasets are scarce, given that embodied CoT is expensive to annotate and often requires dense labeling of multiple perception tasks within a large amount of images.
Consequently, a large-scale embodied CoT dataset and a systematic scaling study is essential to determine whether scaling reasoning supervision yields stronger action generation.

\begin{figure}[!t]
    \centering
    \includegraphics[width=1.0\textwidth]{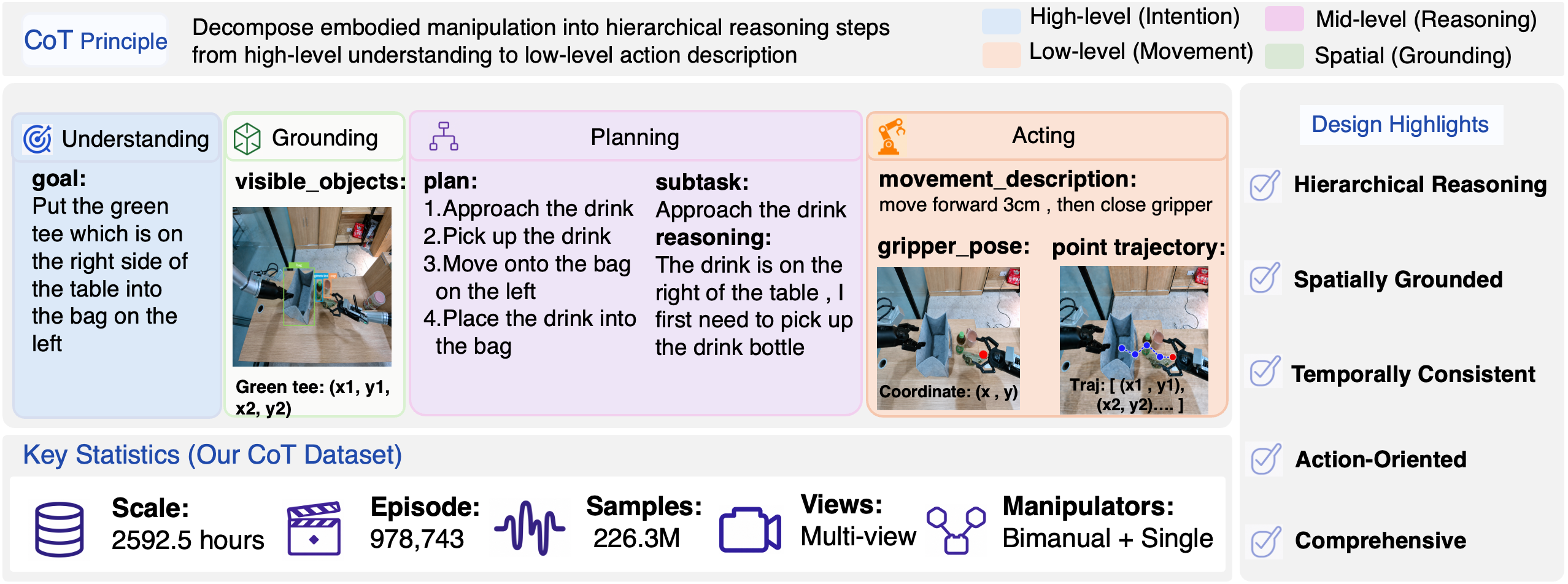}
    \caption{
\textbf{Embodied CoT data format and statistics.}
We introduce a hierarchical embodied CoT schema that decomposes robot manipulation into task understanding, planning, spatial grounding, and action-level descriptions. 
Each sample connects language reasoning with multi-view observations and executable cues. 
Using this schema, we construct a large-scale public embodied CoT dataset covering 2,592.5 hours, 978,743 episodes, and 226.3M samples.
}
    \label{fig:dataset}
\end{figure}

In this paper, we revisit embodied CoT for robot control on an unprecedented scale, with a focus on \emph{linguistic CoT} (see Figure.~\ref{fig:overview}).
We first establishes a comprehensive CoT format and an auto-labeling pipeline, extending prior single-view reasoning to multi-view settings.
Based on this pipeline, we curate the largest embodied CoT robot dataset to date, consisting of 978,743 trajectories, 226.3M samples, and 2592.5 hours of data (see Figure.~\ref{fig:dataset}).
We perform extensive experiments on this dataset in a unified setting to compare the effectiveness of various CoT signals.

We identifies a critical yet overlooked failure mode that we term \textbf{CoT contamination}.
Noisy labels are inevitable in large-scale auto-labeled CoT annotations.
These dense but inaccurate supervision, especially highly-jitterred signals such as unstable bounding boxes and drifting end-effector coordinates across consecutive similar frames, can hinder adaptation to embodied control by imposing inconsistent supervision on semantically similar observations.
We show that this issue can be mitigated through improved annotation pipelines, sparse supervision for unstable grounding tasks, and properly designed reasoning dropout. 
Importantly, reasoning dropout is not merely a regularizer, it acts as training scaffolding that reduces noisy CoT contamination and reshapes internal VLA representations.
This leads to a central conclusion: \textbf{embodied CoT should be treated not merely as a test-time verbalization channel, but as a training signal that reshapes the representation space from perception to action.}

Building on these insights, we present \textbf{ERVLA}, an \textbf{E}mbodied \textbf{R}easoning \textbf{V}ision-\textbf{L}anguage-\textbf{A}ction model trained on a large-scale robot dataset with embodied CoT.
ERVLA leverages a mixture-of-transformers architecture~\cite{black2026pi0visionlanguageactionflowmodel,intelligence2025pi05visionlanguageactionmodelopenworld,tang2025mindhandpurposefulrobotic}, where a vision-language model~\cite{bai2025qwen3vltechnicalreport} processes visual observations and text instructions, and a diffusion transformer (DiT)~\cite{peebles2023scalablediffusionmodelstransformers} generates continuous robot actions via flow matching~\cite{lipman2023flowmatchinggenerativemodeling}.
The DiT is conditioned on the robot state and the key-value (KV) cache from the VLM, allowing action generation to directly access semantic representations produced by the reasoning model.
Drawing inspiration from~\cite{cai2026xiaomi,intelligence2025pi05visionlanguageactionmodelopenworld}, we train the VLM with an auxiliary action-regression loss through query tokens, so that the backbone is guided not only to produce textual reasoning but also to understand action-relevant semantics. This auxiliary supervision stabilizes training, accelerates convergence, and encourages the VLM representation to align high-level reasoning with low-level action generation.
We empirically found that it is important to truncate the KV cache of these tokens in the condition of the flow matching to avoid shortcut copying.

Besides action generation, we supervise the VLM backbone with next-token prediction over embodied CoT traces, bridging semantic reasoning and robotic control.
This supervision serves to reshape the representation space of the VLM for action generation.
During training, we apply random dropout to the reasoning supervision, utilizing an explicit toggle signal to indicate whether to activate reasoning or not as in large language models~\cite{bai2025qwen3vltechnicalreport}.
This not only allows the model to generate actions without reasoning at inference time, but also circumvents the CoT contamination problem described above.

We perform extensive experiments on both simulation benchmarks and real-robot platforms.
Despite trained on LIBERO data~\cite{liu2023liberobenchmarkingknowledgetransfer}, ERVLA achieves a state-of-the-art performance of an 86.9\% average success on LIBERO-Plus~\cite{fei2025liberoplusindepthrobustnessanalysis} and reaches 100\% success on the background and lighting variations of its Spatial track, demonstrating strong zero-shot generalization capabilities.
In VLABench~\cite{zhang2024vlabenchlargescalebenchmarklanguageconditioned}, ERVLA achieves an average success rate of 53.2\%, showcasing superior performance in challenging out-of-distribution settings demanding semantic understanding and instruction following.
In real-robot experiments, ERVLA surpasses competitive state-of-the-art baselines, especially in tackling tasks with semantic ambiguity and long-horizon tasks. The full dataset, model checkpoints, and code are available on our
\href{https://taoshuaiz.github.io/ERVLA/}{project website}.

\section{Methodology}
\label{method}

We start by systematically reviewing and organizing existing linguistic embodied CoT, grouping its major forms and defining the key fields in Sec.~\ref{sec:cot_data}. A detailed review is provided in Appendix~\ref{appendix:related_work}. Second, we introduce ERVLA and study how embodied reasoning should guide action generation in modern VLA architectures built on VLM backbones and diffusion-based action heads in Sec.~\ref{sec:ervla_arch}. Third, we present the training recipe for building a strong reasoning-aware VLA model in Sec.~\ref{sec:training_recipe}.

\subsection{Embodied CoT Data Construction}
\label{sec:cot_data}

We construct a large-scale embodied CoT corpus as the reasoning pre-training source for ERVLA. Rather than treating CoT as free-form explanations attached to trajectories, our pipeline decomposes embodied reasoning into structured categories that correspond to different roles in control: task understanding, spatial grounding, subgoal planning, and action-oriented motion description. This design is intended to make reasoning supervision both interpretable and learnable: high-level fields provide semantic intent, grounded fields align language with visual entities, planning fields expose task progress, and acting fields connect reasoning to executable motion (see Figure.~\ref{fig:dataset}). The corpus is built on open VLA datasets including Bridge~\cite{walke2024bridgedatav2datasetrobot}, Fractal~\cite{zitkovich2023rt}, Droid~\cite{khazatsky2025droidlargescaleinthewildrobot}, MolmoAct~\cite{lee2025molmoactactionreasoningmodels}, and AgiBot~\cite{agibotworldcontributors2025agibotworldcolosseolargescale}, covering both single-arm and dual-arm manipulation, as well as single-view and multi-view observations. In this way, the dataset provides not only language supervision, but a scalable interface for studying how semantic, spatial, planning, and action-level reasoning jointly shape model representations. Detailed annotation procedures, filtering rules, and dataset construction pipelines are provided in Appendix~\ref{appendix:data_pipeline}.

\begin{figure}[!t]
    \centering
    \includegraphics[width=1.0\textwidth]{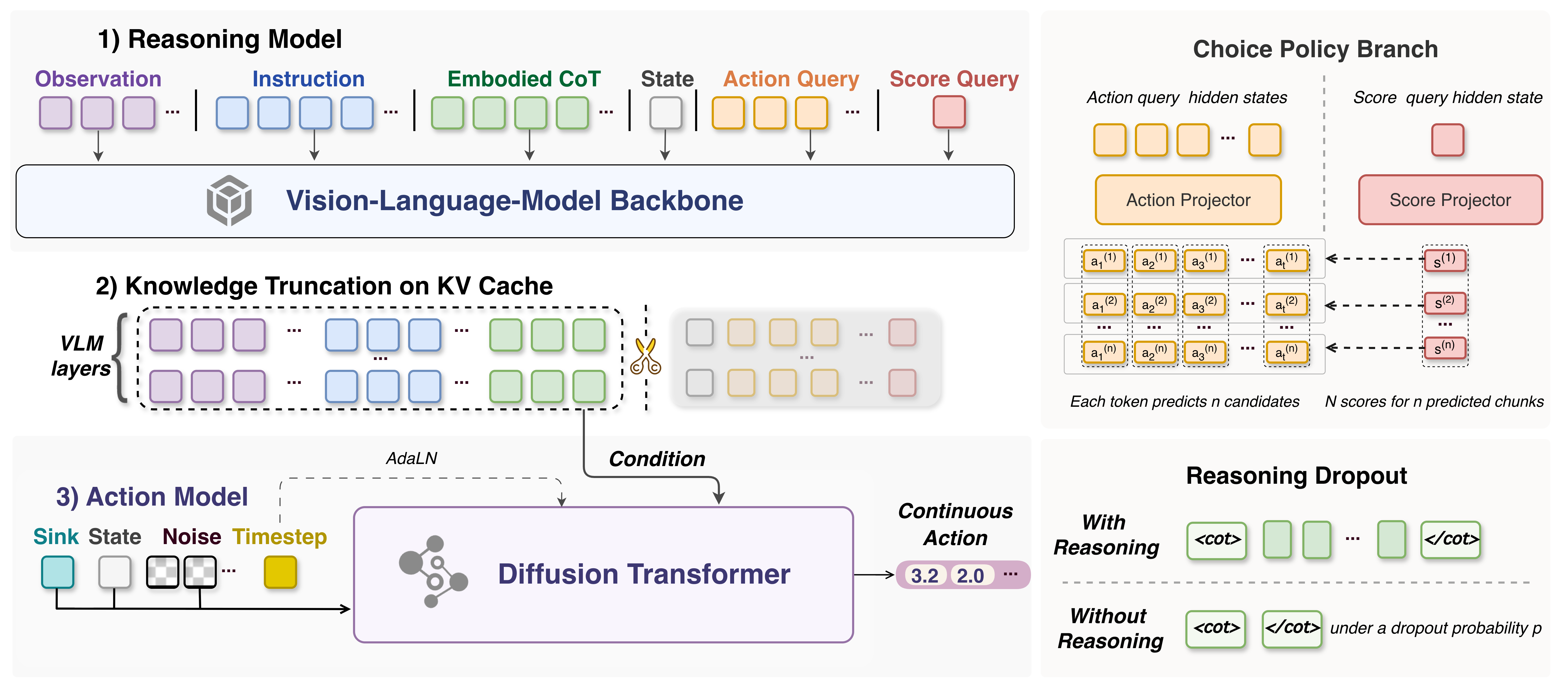}
\caption{
\textbf{Demonstration of ERVLA architecture.}
ERVLA integrates explicit embodied CoT supervision into the VLM backbone, uses auxiliary action-query regression to align semantic reasoning with action, applies knowledge truncation so that the action model attends only to the semantic-prefix KV cache, and generates continuous actions through a diffusion transformer with flow-matching.
}
    \label{fig:method}
\end{figure}

\subsection{ERVLA Architecture}
\label{sec:ervla_arch}

With the structured ECoT corpus established, we next present the ERVLA architecture that integrates reasoning pre-training with end-to-end action generation. ERVLA contains a vision-language reasoning module and a diffusion-based action module as depicted in Figure.~\ref{fig:method}. We instantiate the reasoning backbone with Qwen3-VL-4B~\cite{bai2025qwen3vltechnicalreport} and keep explicit embodied CoT supervision in the standard language-modeling form. Given images $I$, instruction $x$, optional CoT $c$, state $s$, action-query tokens $\{a_i\}$, and a score-query token $a_{\mathrm{score}}$, the VLM outputs hidden states and per-layer key-value caches:
\begin{equation}
\mathbf{H}^{\mathrm{vlm}}, \{(\mathbf{K}^{\mathrm{vlm}}_{\ell}, \mathbf{V}^{\mathrm{vlm}}_{\ell})\}_{\ell=1}^{L}
= f_{\mathrm{vlm}}(I, x, c, s, \{a_i\}, a_{\mathrm{score}}).
\end{equation}
The input includes trainable control interfaces: \texttt{<state>} token is replaced by projected state embedding, \texttt{<a\_i>} by action-query embeddings, and \texttt{<score>} by a score-query embedding. Unlike ThinkAct-style designs~\cite{huang2025thinkactvisionlanguageactionreasoningreinforced}, which mainly use the VLM for planning while modeling actions outside the backbone, ERVLA injects action-sensitive supervision directly into the VLM. We also avoid compressed action tokenization such as FAST~\cite{pertsch2025fastefficientactiontokenization}, which introduces a separate action-token space and may cause discretization error. Instead, ERVLA preserves CoT in the native language space and uses an auxiliary \emph{choice policy}~\cite{qi2025coordinatedhumanoidmanipulationchoice} branch to predict $N$ candidate action chunks and their scores:
\begin{equation}
\hat{\mathbf{a}}_{t}^{(n)} = g_{\mathrm{act}}(\mathbf{H}_{a})_{t,n},
\qquad
t=1,\ldots,T,\;\; n=1,\ldots,N,
\qquad
\hat{\mathbf{r}} = g_{\mathrm{score}}(\mathbf{H}_{s}).
\end{equation}
Here, $\mathbf{H}_{a}$ and $\mathbf{H}_{s}$ denote hidden states at the \texttt{<a\_i>} and \texttt{<score>} positions. Each \texttt{<a\_i>} corresponds to one temporal slot, where the action head predicts $N$ vectors $\hat{\mathbf{a}}_{t}^{(n)} \in \mathbb{R}^{D}$. Grouping vectors with the same candidate index forms $N$ continuous chunks, while $\hat{\mathbf{r}} \in \mathbb{R}^{N}$ scores them. The final action is generated by a DiT-style diffusion policy rather than next-token prediction:
\begin{equation}
\hat{\mathbf{A}}^{(n)}
=
[\hat{\mathbf{a}}_{1}^{(n)}, \ldots, \hat{\mathbf{a}}_{T}^{(n)}]
\in \mathbb{R}^{T \times D},
\qquad
\hat{\mathbf{v}}_{\theta}
=
f_{\mathrm{dit}}(\mathbf{z}_{\tau}, \tau, s \mid \{(\mathbf{K}^{\mathrm{vlm}}_{\ell}, \mathbf{V}^{\mathrm{vlm}}_{\ell})\}) .
\end{equation}
Here, $\mathbf{z}_{\tau}$ is the noisy action, and the DiT input consists of a sink token, projected state tokens, and projected noisy action tokens. The diffusion timestep $\tau$ is encoded by an MLP and injected into each DiT block through AdaLN conditioning~\cite{peebles2023scalablediffusionmodelstransformers}. Rather than using pooled VLM features, DiT directly attends to the VLM's per-layer key-value cache. 

We further introduce \emph{knowledge truncation}: although the VLM receives both the semantic prefix and appended control-query turns during training, DiT conditions only on the semantic-prefix cache:
\begin{equation}
\small
\mathrm{Attn}(\mathbf{Q}, [\mathbf{K}^{\mathrm{KT}}_{\ell}; \mathbf{K}^{\mathrm{dit}}_{\ell}], [\mathbf{V}^{\mathrm{KT}}_{\ell}; \mathbf{V}^{\mathrm{dit}}_{\ell}]),
\quad
\{(\mathbf{K}^{\mathrm{KT}}_{\ell}, \mathbf{V}^{\mathrm{KT}}_{\ell})\}
=
\mathrm{SlicePrefix}\big(
\{(\mathbf{K}^{\mathrm{vlm}}_{\ell}, \mathbf{V}^{\mathrm{vlm}}_{\ell})\}, m_{\mathrm{cond}}
\big).
\end{equation}
where $m_{\mathrm{cond}}$ excludes appended state and action-query turns. In this way, DiT reads only the semantic memory before the control-query tokens, preventing it from exploiting synthetic query shortcuts and reducing interference from training-specific token patterns.

\subsection{Training Recipe}
\label{sec:training_recipe}

Unlike knowledge-insulated VLA training~\cite{driess2025knowledgeinsulatingvisionlanguageactionmodels}, which blocks gradients from the continuous action expert to the VLM backbone, ERVLA enables the flow-matching loss from the DiT action head to back-propagate into the VLM backbone, making the whole reasoning-action system fully end-to-end trainable. Combining explicit CoT verbalization, reasoning dropout, choice policy prediction, and knowledge truncation, ERVLA enables action-sensitive VLM representations, cleaner semantic memory for DiT, and natural support for test-time reasoning dropout, where CoT can be partially used, sparsely refreshed, or omitted.

We optimize ERVLA with the objective $\mathcal{L}=\lambda_{\mathrm{vlm}}\mathcal{L}_{\mathrm{vlm}}+\lambda_{\mathrm{flow}}\mathcal{L}_{\mathrm{flow}}+\lambda_{\mathrm{choice}}\mathcal{L}_{\mathrm{choice}}+\lambda_{\mathrm{score}}\mathcal{L}_{\mathrm{score}}$. Here, $\mathcal{L}_{\mathrm{vlm}}$ is the token-level CoT cross-entropy loss, and $\mathcal{L}_{\mathrm{flow}}$ is the rectified flow loss for continuous actions. For the choice policy branch, the model predicts $N$ candidate chunks and supervises the best one under mean absolute error:
\begin{equation}
\small
\mathcal{L}_{\mathrm{choice}}
=
\frac{1}{B}\sum_{b=1}^{B}\min_{n} d_b^{(n)},
\qquad
d_b^{(n)}
=
\frac{1}{T_bD}
\left\|
\hat{\mathbf{A}}_{b}^{(n)}-\mathbf{A}_{b}^{*}
\right\|_{1}.
\end{equation}
The score branch predicts these candidate-wise errors:
\begin{equation}
\small
\mathcal{L}_{\mathrm{score}}
=
\frac{1}{B}
\sum_{b=1}^{B}
\left\|
\hat{\mathbf r}_b
-
\mathrm{sg}\!\left([d_b^{(1)},\dots,d_b^{(N)}]\right)
\right\|_2^2 ,
\end{equation}
where $\hat{\mathbf r}_b \in \mathbb{R}^{N}$ is the predicted scores, and $\mathrm{sg}(\cdot)$ denotes stop-gradient through the target errors. 

We apply \emph{reasoning dropout}~\cite{chen2025trainingstrategiesefficientembodied} during training: each sample is converted to either \texttt{/cot} or \texttt{/no\_cot} with probability $p_{\mathrm{cot}}$. Thus, explicit CoT is an optional training condition rather than a mandatory inference trace. This reduces reliance on visible reasoning, mitigates noisy CoT supervision, encourages reasoning to be internalized into backbone states and the cache-conditioned action interface, and enables test-time reasoning dropout while preserving the benefits of reasoning-aware pre-training.

\section{Experiment}
\label{experiment}

We organize our experiments around three questions:
\begin{itemize}[leftmargin=*, nosep]
    \item \emph{What kinds of embodied CoT are more useful for action policy learning?}
    
    \item \emph{Can embodied CoT provide an effective transfer interface that turns stronger VLM backbones into stronger VLA action policies?}
    
    \item \emph{Can embodied CoT serve as a scalable pre-training signal for generalizable action generation?}

\end{itemize}

To answer them, we first conduct controlled studies with a unified backbone and autoregressive prediction to identify useful CoT fields, evaluate their pre-training value, and examine VLM-to-VLA transfer with ECoT in Sec.~\ref{sec:explore}. We then compare ERVLA with strong baselines and ablate reasoning--action integration and CoT pre-training scale in Sec.~\ref{sec:study}. Finally, we evaluate real-world robot tasks requiring semantic understanding, robust scene grounding, and long-horizon capabilities in Sec.~\ref{sec:real-world}.

\paragraph{Benchmarks.}
We evaluate on three complementary benchmarks:
\begin{itemize}[leftmargin=*, nosep]
    \item \textbf{LIBERO}~\cite{liu2023liberobenchmarkingknowledgetransfer} is one of the most widely used manipulation benchmarks and provides a strong testbed for studies, but its relatively fixed setting can encourage benchmark-specific overfitting.

    \item \textbf{LIBERO-Plus}~\cite{fei2025liberoplusindepthrobustnessanalysis} extends LIBERO with camera, state, language, background, and layout shifts, providing a stronger test of zero-shot transfer and robustness beyond the original distribution.

    \item \textbf{VLABench}~\cite{zhang2024vlabenchlargescalebenchmarklanguageconditioned} is a harder benchmark with one in-distribution track and multiple generalization tracks, covering cross-category transfer, commonsense reasoning, semantic instruction following, and unseen texture variation. It is suited for testing semantic and distribution shifts.
\end{itemize}

\vspace{-3mm}

\paragraph{Baselines.}
On LIBERO-Plus, we compare with ECoT~\cite{zawalski2024robotic}, Emma-X~\cite{sun2024emmaxembodiedmultimodalaction}, OpenVLA-OFT~\cite{kim2025finetuningvisionlanguageactionmodelsoptimizing}, UniVLA~\cite{bu2025univlalearningacttaskcentric}, WorldVLA~\cite{cen2025worldvlaautoregressiveactionworld}, $\pi_0$~\cite{black2026pi0visionlanguageactionflowmodel}, $\pi_0$-FAST~\cite{pertsch2025fastefficientactiontokenization}, Spatial Forcing~\cite{li2025spatialforcingimplicitspatial}, PokeVLA~\cite{zheng2026pokevlaempoweringpocketsizedvisionlanguageaction}, and $\pi_{0.5}$~\cite{intelligence2025pi05visionlanguageactionmodelopenworld}. On VLABench, we additionally compare with X-VLA~\cite{zheng2025xvlasoftpromptedtransformerscalable} and ACoT~\cite{zhong2026acotvlaactionchainofthoughtvisionlanguageaction}. The works cover linguistic, visual, and latent reasoning. We report CoT scaling variants of ERVLA and ablations of the choice policy and knowledge truncation. Full training details are provided in Appendix~\ref{appendix:training_details}.

\vspace{-3mm}

\paragraph{VLM backbones.}
To examine whether stronger VLMs transfer better to action through embodied CoT, we evaluate diverse backbones spanning a broad range on the OpenVLM Leaderboard~\cite{duan2024vlmevalkit}: PaliGemma-2-3B~\cite{beyer2024paligemmaversatile3bvlm}, Florence-2-large~\cite{xiao2023florence2advancingunifiedrepresentation}, Cosmos-Reason2-2B~\cite{nvidia2025cosmosreason1physicalcommonsense}, Gemma-4-E2B-it, Qwen2.5-VL-3B/7B~\cite{qwen2025qwen25technicalreport}, and Qwen3-VL-2B/4B/8B~\cite{bai2025qwen3vltechnicalreport}. All models share the same action-token interface and are evaluated under the StarVLA framework~\cite{community2026starvlalegolikecodebasevisionlanguageaction}, with training details provided in Appendix~\ref{appendix:training_details}.


\definecolor{headergray}{RGB}{242,244,247}
\definecolor{descblue}{RGB}{232,240,254}
\definecolor{coordorange}{RGB}{255,243,224}
\definecolor{resultgray}{RGB}{248,249,250}
\definecolor{dropgreen}{RGB}{232,245,233}
\definecolor{fullviolet}{RGB}{243,235,255}

\begin{table*}[!t]
\centering
\scriptsize
\setlength{\tabcolsep}{4.0pt}
\renewcommand{\arraystretch}{1.14}
\caption{\textbf{Ablation of ECoT fields on average success rate in VLABench~\cite{zhang2024vlabenchlargescalebenchmarklanguageconditioned}.} The lower block reports performance changes with and without Bridge pre-training~\cite{walke2024bridgedatav2datasetrobot}, relative to the first column.}
\label{tab:cottype}

\resizebox{\textwidth}{!}{%
\begin{tabular}{>{\raggedright\arraybackslash}p{1.45cm} >{\raggedright\arraybackslash}p{1.55cm} *{15}{c}}
\toprule
\rowcolor{headergray}
\multicolumn{2}{c}{\textbf{CoT Field}}
& \multicolumn{15}{c}{\textbf{Ablation Setting}} \\
\cmidrule(lr){1-2} \cmidrule(lr){3-17}

\multirow{5}{*}{\centering\shortstack{\textit{Textual}\\\textit{Description}}}
& \cellcolor{descblue}\textbf{Goal}
&   & \checkmark &   &   &   &   &   &   &   & \checkmark &   &   &   &   & \checkmark \\

& \cellcolor{descblue}\textbf{Planning}
&   &   & \checkmark &   &   &   &   &   &   & \checkmark &   &   &   &   & \checkmark \\

& \cellcolor{descblue}\textbf{Subtask}
&   &   &   & \checkmark &   &   &   &   &   & \checkmark &   & \checkmark &   & \checkmark & \checkmark \\

& \cellcolor{descblue}\textbf{Movement}
&   &   &   &   & \checkmark &   &   &   &   & \checkmark &   &   & \checkmark & \checkmark & \checkmark \\

& \cellcolor{descblue}\textbf{Reasoning}
&   &   &   &   &   & \checkmark &   &   &   & \checkmark &   & \checkmark & \checkmark &   & \checkmark \\

\midrule

\multirow{3}{*}{\centering\shortstack{\textit{Numerical}\\\textit{Coordinate}}}
& \cellcolor{coordorange}\textbf{Gripper}
&   &   &   &   &   &   & \checkmark &   &   &   & \checkmark &   &   &   & \checkmark \\

& \cellcolor{coordorange}\textbf{Point traj.}
&   &   &   &   &   &   &   & \checkmark &   &   & \checkmark &   &   & \checkmark & \checkmark \\

& \cellcolor{coordorange}\textbf{Bounding box}
&   &   &   &   &   &   &   &   & \checkmark &   & \checkmark &   &   &   & \checkmark \\

\midrule

\rowcolor{resultgray}
\multicolumn{2}{l}{\textbf{w/o Pretraining}}
& \textbf{19.0} & \negv{1.2} & \negv{0.8} & \negv{0.6} & \pos{4.1} & \negv{1.0} & \negv{0.7} & \pos{4.8} & \negv{1.4} & \pos{4.9} & \pos{2.7} & \negv{1.3} & \pos{5.2} & \pos{7.4} & \cellcolor{fullviolet}\pos{8.2} \\

\rowcolor{resultgray}
\multicolumn{2}{l}{\textbf{w/ Pretraining}}
& \textbf{25.2} & \negv{0.8} & \negv{0.5} & \negv{0.7} & \pos{2.0} & \negv{0.9} & \negv{5.6} & \pos{1.4} & \negv{6.1} & \pos{3.1} & \negv{2.4} & \negv{1.6} & \pos{3.0} & \pos{3.7} & \cellcolor{fullviolet}\pos{2.5} \\

\rowcolor{resultgray}
\multicolumn{2}{l}{\textbf{w/ Pretraining + Dropout}}
& \textbf{25.2} & \negv{0.6} & \negv{0.3} & \negv{0.5} & \pos{1.9} & \negv{0.6} & \negv{0.8} & \pos{3.0} & \negv{1.0} & \pos{2.9} & \pos{1.5} & \negv{0.9} & \pos{3.2} & \pos{4.4} & \cellcolor{fullviolet}\pos{4.0} \\

\bottomrule
\end{tabular}%
}
\end{table*}

\subsection{Exploring Effective Embodied Chain-of-Thought Signals}
\label{sec:explore}

To study individual and combined embodied CoT signals, we annotate VLABench~\cite{zhang2024vlabenchlargescalebenchmarklanguageconditioned} and LIBERO~\cite{liu2023liberobenchmarkingknowledgetransfer} using the format in Sec.~\ref{sec:cot_data}. We test on Qwen3-VL-4B~\cite{bai2025qwen3vltechnicalreport} with a fixed autoregressive CoT+FAST~\cite{pertsch2025fastefficientactiontokenization} action interface, so performance differences mainly reflect the CoT signal itself.

\noindent\textbf{Action-related CoT signals support action generation more directly than high-level understanding alone.}
As shown in Table~\ref{tab:cottype}, isolated high-level fields do not help action learning under direct training: \textit{Goal}, \textit{Planning}, \textit{Subtask}, and \textit{Reasoning} all lead to small drops, with \textit{Goal} decreasing performance by -1.2. In contrast, action-related fields are more effective: \textit{Movement} improves performance by +4.1 and \textit{Point trajectory} by +4.8, while coordinate-only cues such as \textit{Bounding box} remain insufficient when used alone. High-level understanding becomes useful only when coupled with concrete action guidance, as \textit{Movement+Reasoning} reaches +5.2 and \textit{Subtask+Movement+Point trajectory} reaches +7.4. Full ECoT gives the strongest gain, suggesting that effective embodied CoT should connect semantic understanding to executable motion rather than merely describe task intent.

\noindent\textbf{Pretraining helps only with reliable CoT supervision, while reasoning dropout mitigates CoT contamination.}
With ECoT pretraining on Bridge~\cite{walke2024bridgedatav2datasetrobot}, unreliable coordinate fields become harmful: \textit{Gripper} and \textit{Bounding box} cause the largest drops, reducing performance by -5.6 and -6.1, respectively. Unlike benchmark annotations obtained by simulator replay and thus closer to ground truth, large-scale detector-generated labels are lossy and can impose inconsistent supervision on semantically similar frames. Reasoning dropout substantially reduces this damage, shrinking the drops of \textit{Gripper} and \textit{Bounding box} to -0.8 and -1.0, while preserving useful action-oriented supervision such as \textit{Point trajectory}, which improves from +1.4 without dropout to +3.0 with dropout. Overall, CoT pretraining works best when reliable acting signals are retained and unstable grounding fields are regularized rather than densely imposed.

\noindent\textbf{Explicit CoT does not automatically scale under autoregressive action modeling.}
With full CoT supervision, Bridge-only pretraining provides modest gains, but Table~\ref{tab:badpretrain} shows that simply adding more CoT data does not reliably improve an autoregressive CoT+action-token model. As the pretraining mixture expands to Fractal~\cite{zitkovich2023rt}, MolmoAct~\cite{lee2025molmoactactionreasoningmodels}, and Droid~\cite{khazatsky2025droidlargescaleinthewildrobot}, performance gradually degrades across VLABench tracks. In the four-source setting, the drops reach -3.6 on In-dist., -3.0 on Cross-category, and -3.4 on Texture, indicating that more explicit CoT can introduce more prefix noise rather than better control. We attribute this to the fragility of autoregressive action decoding under dense and drifting supervision: reasoning traces vary in length and quality, grounding noise perturbs the action prefix, and CoT errors directly affect subsequent action tokens. Thus, explicit CoT is useful as supervision, but autoregressive action decoding is a poor substrate for scaling it.

\noindent\textbf{Embodied CoT makes stronger VLMs more reliably transferable to stronger VLAs.}
Figure.~\ref{fig:ecot_scaling_ablation} left shows that under non-CoT recipe, backbone strength and downstream performance remain weakly correlated. With explicit embodied CoT, stronger VLMs become more consistently effective: the best results are dominated by Qwen3-VL series~\cite{bai2025qwen3vltechnicalreport}, with Qwen3-VL-4B and leading on LIBERO \emph{Spatial} and \emph{Goal} and VLABench \emph{In-dist.}, \emph{Category}, and \emph{Common} as detailed in Table~\ref{tab:libero_liberoplus_combined} of Appendix~\ref{appendix: detailed experiments}. This suggests that embodied CoT is not merely an auxiliary label space, but a transfer interface that converts stronger VLM semantic priors into action-relevant representations.



\begin{table*}[!t]
\centering
\scriptsize
\setlength{\tabcolsep}{5.2pt}
\renewcommand{\arraystretch}{1.10}
\caption{\textbf{Comparison of scaling embodied CoT pre-training data on LIBERO~\cite{liu2023liberobenchmarkingknowledgetransfer} and VLABench~\cite{zhang2024vlabenchlargescalebenchmarklanguageconditioned}.} All settings use FAST~\cite{pertsch2025fastefficientactiontokenization} action representation with embodied CoT supervision and reasoning dropout. Values below the first row report performance changes relative to it.}
\label{tab:badpretrain}

\resizebox{\textwidth}{!}{%
\begin{tabular}{lccccccccc}
\toprule
\multirow{2}{*}{\textbf{Pre-Training Data (with dropout)}}
& \multicolumn{4}{c}{\textbf{LIBERO}}
& \multicolumn{5}{c}{\textbf{VLABench}} \\
\cmidrule(lr){2-5} \cmidrule(lr){6-10}
& \textbf{Spatial}
& \textbf{Object}
& \textbf{Goal}
& \textbf{Long}
& \textbf{In-dist.}
& \textbf{Category}
& \textbf{Common}
& \textbf{Instruction}
& \textbf{Texture} \\
\midrule

--
& 96.4 & 96.6 & 91.8 & 88.6
& 38.0 & 29.0 & 17.6 & 25.8 & 30.0 \\

Bri.~\cite{walke2024bridgedatav2datasetrobot}
& \pos{0.4} & \pos{0.8} & \pos{0.4} & \pos{0.6}
& \pos{1.0} & \pos{0.8} & \pos{0.6} & \negv{0.4} & \pos{1.2} \\

Bri.~\cite{walke2024bridgedatav2datasetrobot}+Fra.~\cite{zitkovich2023rt}
& \pos{0.2} & \negv{0.6} & \pos{0.2} & \negv{0.8}
& \negv{1.8} & \pos{0.2} & \pos{0.4} & \pos{0.2} & \negv{1.0} \\

Bri.~\cite{walke2024bridgedatav2datasetrobot}+Fra.~\cite{zitkovich2023rt}+Mol.~\cite{lee2025molmoactactionreasoningmodels}
& \negv{0.8} & \negv{1.0} & \negv{0.6} & \negv{1.4}
& \negv{1.2} & \negv{1.0} & \negv{0.8} & \negv{1.4} & \negv{1.2} \\

Bri.~\cite{walke2024bridgedatav2datasetrobot}+Fra.~\cite{zitkovich2023rt}+Mol.~\cite{lee2025molmoactactionreasoningmodels}+Dro.~\cite{khazatsky2025droidlargescaleinthewildrobot}
& \negv{1.8} & \negv{1.6} & \negv{2.0} & \negv{2.4}
& \negv{3.6} & \negv{3.0} & \negv{2.2} & \negv{3.2} & \negv{3.4} \\

\bottomrule

\vspace{-5mm}

\end{tabular}%
}
\end{table*}

\begin{table*}[!t]
\centering
\scriptsize
\setlength{\tabcolsep}{4.2pt}
\renewcommand{\arraystretch}{1.08}
\caption{
\textbf{Quantitative comparison on the LIBERO-Plus~\cite{fei2025liberoplusindepthrobustnessanalysis} benchmark.}
Models are post-trained on LIBERO~\cite{liu2023liberobenchmarkingknowledgetransfer} and evaluated by zero-shot direct transfer.
Gray-shaded rows are ERVLA ablations: \textit{No Choice (E2E)} allows DiT loss back-propagated to the VLM; \textit{No CoT} removes explicit embodied chain-of-thought supervision while keeping the same action-learning pipeline; \textit{No Choice + KI} blocks this back-propagation, resembling ThinkAct-style~\cite{huang2025thinkactvisionlanguageactionreasoningreinforced} planning and action separation; \textit{Choice w/o KT} lets DiT access the CoT+choice cache, which can introduce shortcut learning.
}
\label{tab:libero_plus_comparison}

\resizebox{\textwidth}{!}{%
\begin{tabular}{lcccccccccccc}
\toprule
\multirow{2}{*}{\textbf{Method}}
& \multicolumn{4}{c}{\textbf{Task Suite}}
& \multicolumn{7}{c}{\textbf{Perturbation Type}}
& \multirow{2}{*}{\textbf{Total}} \\
\cmidrule(lr){2-5} \cmidrule(lr){6-12}
& \textbf{Spatial}
& \textbf{Object}
& \textbf{Goal}
& \textbf{Long}
& \textbf{Camera}
& \textbf{Robot}
& \textbf{Language}
& \textbf{Light}
& \textbf{Background}
& \textbf{Noise}
& \textbf{Layout}
& \\
\midrule

ECoT~\cite{zawalski2024robotic}
& 31.8 & 27.9 & 30.6 & 8.6
& 0.3 & 26.8 & 40.2 & 42.6 & 16.4 & 10.2 & 36.9
& 24.3 \\

Emma-X~\cite{sun2024emmaxembodiedmultimodalaction}
& 33.4 & 28.4 & 31.2 & 8.0
& 0.2 & 28.6 & 42.0 & 42.8 & 17.6 & 10.0 & 37.4
& 25.1 \\

OpenVLA-OFT~\cite{kim2025finetuningvisionlanguageactionmodelsoptimizing}
& 84.0 & 66.5 & 63.0 & 66.4
& 56.4 & 31.9 & 79.5 & 88.7 & 93.3 & 75.8 & 74.2
& 69.6 \\

UniVLA~\cite{bu2025univlalearningacttaskcentric}
& 55.5 & 36.7 & 40.7 & 39.9
& 1.8 & 46.2 & 69.6 & 69.0 & 81.0 & 21.2 & 31.9
& 42.9 \\

WorldVLA~\cite{cen2025worldvlaautoregressiveactionworld}
& 32.5 & 28.6 & 31.8 & 8.2
& 0.1 & 27.9 & 41.6 & 43.7 & 17.1 & 10.9 & 38.0
& 25.0 \\

$\pi_0$~\cite{black2026pi0visionlanguageactionflowmodel}
& 60.7 & 61.4 & 44.9 & 48.4
& 13.8 & 6.0 & 58.8 & 85.0 & 81.4 & 79.0 & 68.8
& 53.6 \\

$\pi_0$-FAST~\cite{pertsch2025fastefficientactiontokenization}
& 74.4 & 72.7 & 57.6 & 43.4
& 65.1 & 21.6 & 61.0 & 73.2 & 73.2 & 74.4 & 68.8
& 61.6 \\

Spatial Forcing~\cite{li2025spatialforcingimplicitspatial}
& 52.9 & 31.0 & 28.2 & 5.4
& 20.1 & 13.4 & 40.9 & 29.1 & 33.4 & 25.7 & 39.3
& 29.1 \\

PokeVLA~\cite{zheng2026pokevlaempoweringpocketsizedvisionlanguageaction}
& 85.4 & 81.8 & 77.6 & 72.7
& 84.7 & 46.1 & 84.8 & 94.6 & 82.6 & 89.8 & 77.2
& 79.3 \\

$\pi_{0.5}$~\cite{intelligence2025pi05visionlanguageactionmodelopenworld}
& 90.4 & \textbf{89.9} & \textbf{81.0} & 80.8
& 71.7 & 75.5 & 85.9 & \textbf{96.1} & \textbf{95.7} & 86.4 & \textbf{87.5}
& 85.5 \\

\midrule

\rowcolor{gray!12}
No Choice 
& 70.8 & 65.4 & 58.6 & 55.2
& 54.6 & 42.8 & 64.2 & 72.4 & 66.0 & 68.8 & 62.0
& 61.9 \\

\rowcolor{gray!12}
No CoT
& 77.4 & 71.8 & 65.2 & 62.0
& 68.6 & 43.2 & 72.6 & 82.0 & 72.6 & 76.4 & 68.8
& 70.8 \\

\rowcolor{gray!12}
No Choice + Knowledge Insulation
& 83.8 & 78.6 & 71.4 & 69.0
& 80.4 & 43.8 & 81.0 & 91.6 & 79.2 & 85.4 & 74.6
& 76.5 \\

\rowcolor{gray!12}
Choice + No Knowledge Truncation
& 89.2 & 88.6 & 79.4 & 79.8
& 70.8 & 73.4 & 84.6 & 94.2 & 94.4 & 85.6 & 86.2
& 84.7 \\

\midrule

\rowcolor{blue!5}
\textbf{ERVLA (Ours)}
& \textbf{96.2} & 89.6 & 79.6 & \textbf{82.1}
& \textbf{77.2} & 75.3 & \textbf{87.1} & 95.1 & 94.7 & \textbf{92.3} & 86.4
& \textbf{86.9} \\

\bottomrule
\end{tabular}%
}

\end{table*}

\begin{table*}[!t]
\centering
\scriptsize
\setlength{\tabcolsep}{3.4pt}
\renewcommand{\arraystretch}{1.08}
\caption{\textbf{Comparison on the VLABench benchmark.} SR, PS, and IS denote Success Rate, Progress Score, and Intention Score, respectively.}
\label{tab:vlabench_comparison}

\resizebox{\textwidth}{!}{%
\begin{tabular}{lcccccccccccccccccc}
\toprule
\multirow{2}{*}{\textbf{Method}}
& \multicolumn{3}{c}{\textbf{In-dist.}}
& \multicolumn{3}{c}{\textbf{Cross Category}}
& \multicolumn{3}{c}{\textbf{Commonsense}}
& \multicolumn{3}{c}{\textbf{Instruction}}
& \multicolumn{3}{c}{\textbf{Texture}}
& \multicolumn{3}{c}{\textbf{Avg.}} \\
\cmidrule(lr){2-4}
\cmidrule(lr){5-7}
\cmidrule(lr){8-10}
\cmidrule(lr){11-13}
\cmidrule(lr){14-16}
\cmidrule(lr){17-19}
& SR $\uparrow$ & PS $\uparrow$ & IS $\uparrow$
& SR $\uparrow$ & PS $\uparrow$ & IS $\uparrow$
& SR $\uparrow$ & PS $\uparrow$ & IS $\uparrow$
& SR $\uparrow$ & PS $\uparrow$ & IS $\uparrow$
& SR $\uparrow$ & PS $\uparrow$ & IS $\uparrow$
& SR $\uparrow$ & PS $\uparrow$ & IS $\uparrow$ \\
\midrule

$\pi_0$~\cite{black2026pi0visionlanguageactionflowmodel}
& 47.0 & 62.7 & 67.8
& 21.2 & 33.6 & 44.0
& 29.1 & 43.0 & 54.9
& 17.3 & 38.7 & 58.0
& 32.2 & 42.5 & 50.6
& 29.4 & 44.1 & 55.0 \\

$\pi_0$-FAST~\cite{pertsch2025fastefficientactiontokenization}
& 56.2 & 66.8 & 72.4
& 31.0 & 38.2 & 47.8
& 38.0 & 48.6 & 56.8
& 35.0 & 45.0 & 59.4
& 39.0 & 49.0 & 56.8
& 39.8 & 49.5 & 58.6 \\

X-VLA~\cite{zheng2025xvlasoftpromptedtransformerscalable}
& -- & 67.8 & --
& -- & 25.1 & --
& -- & 48.2 & --
& -- & 63.1 & --
& -- & -- & --
& -- & 51.1 & -- \\

ACoT-VLA~\cite{zhong2026acotvlaactionchainofthoughtvisionlanguageaction}
& -- & 66.1 & 79.8
& -- & 38.9 & 54.1
& -- & 37.8 & 52.3
& -- & 39.6 & 56.8
& -- & 54.6 & \textbf{74.6}
& -- & 47.4 & 63.5 \\

$\pi_{0.5}$~\cite{intelligence2025pi05visionlanguageactionmodelopenworld}
& 65.4 & 77.8 & 80.4
& 38.2 & 49.7 & 52.0
& 43.9 & \textbf{57.3} & \textbf{60.0}
& 48.2 & 64.2 & 67.0
& 44.9 & \textbf{62.3} & 65.0
& 48.1 & 62.3 & 64.9 \\

\midrule

\rowcolor{gray!12}
No Choice 
& 50.2 & 57.8 & 65.0
& 33.6 & 42.0 & 49.6
& 36.4 & 45.4 & 52.8
& 42.4 & 54.6 & 61.2
& 33.4 & 41.2 & 50.4
& 39.2 & 48.2 & 55.8 \\

\rowcolor{gray!12}
No CoT
& 52.6 & 59.6 & 66.0
& 34.8 & 44.8 & 51.6
& 38.8 & 47.8 & 54.0
& 44.2 & 56.4 & 62.4
& 34.0 & 43.4 & 51.6
& 40.9 & 50.4 & 57.1 \\

\rowcolor{gray!12}
No Choice + Knowledge Insulation
& 55.0 & 61.4 & 67.2
& 36.0 & 47.6 & 53.8
& 41.2 & 50.0 & 55.0
& 46.0 & 58.2 & 63.4
& 34.8 & 45.4 & 52.6
& 42.6 & 52.5 & 58.4 \\

\rowcolor{gray!12}
Choice + No Knowledge Truncation
& 62.0 & 73.0 & 76.2
& 42.4 & 55.2 & 59.4
& 43.0 & 52.6 & 56.4
& 53.6 & 66.4 & 70.2
& 35.0 & 51.8 & 54.6
& 47.2 & 59.8 & 63.4 \\

\midrule

\rowcolor{blue!5}
\textbf{ERVLA (Ours)}
& \textbf{69.7} & \textbf{81.1} & \textbf{84.2}
& \textbf{47.0} & \textbf{61.0} & \textbf{66.4}
& \textbf{44.0} & 55.0 & 57.2
& \textbf{58.0} & \textbf{70.2} & \textbf{73.8}
& \textbf{47.4} & \textbf{62.3} & 70.6
& \textbf{53.2} & \textbf{65.9} & \textbf{70.4} \\

\bottomrule
\end{tabular}%
}

\vspace{-3mm}

\end{table*}

\subsection{Studies of ERVLA Design Choices and ECoT Scaling}
\label{sec:study}

We move beyond the controlled LIBERO setting to LIBERO-Plus. Sec.~\ref{sec:explore} motivates our ERVLA design: naive autoregressive for CoT and action generation does not scale reliably, and explicit CoT is most useful when it reshapes action-relevant representations rather than serving as a fragile token prefix. We instantiate ERVLA with Qwen3-VL-4B~\cite{bai2025qwen3vltechnicalreport}, add a \emph{choice policy~\cite{qi2025coordinatedhumanoidmanipulationchoice}} branch for candidate action chunks and scores, and apply \emph{knowledge truncation} so DiT conditions on semantic CoT memory instead of control-query tokens. With full embodied CoT pre-training and reasoning dropout, we obtain the final ERVLA model and analyze both design choices and pre-training scale.

\noindent\textbf{ERVLA achieves state-of-the-art performance through the choice policy and knowledge truncation.}
As shown in Tables~\ref{tab:libero_plus_comparison} and~\ref{tab:vlabench_comparison}, ERVLA reaches 86.9 on LIBERO-Plus, surpassing $\pi_{0.5}$ (85.5), and obtains 53.2/70.4/65.9 average SR/IS/PS on VLABench, outperforming strong baselines including atent-interface-based UniVLA~\cite{bu2025univlalearningacttaskcentric} and visual-prediction-enhanced WorldVLA~\cite{cen2025worldvlaautoregressiveactionworld}. These gains come from the proposed reasoning--action interface rather than simply adding explicit CoT. Removing the choice policy branch weakens action-aware representation shaping, while replacing knowledge truncation with naive end-to-end conditioning allows DiT to exploit synthetic control-query tokens and causes large drops on both benchmarks. These results show that ERVLA benefits from both components: the choice policy injects action-level discrimination into the VLM, and knowledge truncation provides DiT with cleaner semantic memory for continuous-action generation.

\begin{wrapfigure}{r}{0.48\textwidth}
    \centering
    \includegraphics[width=0.46\textwidth]{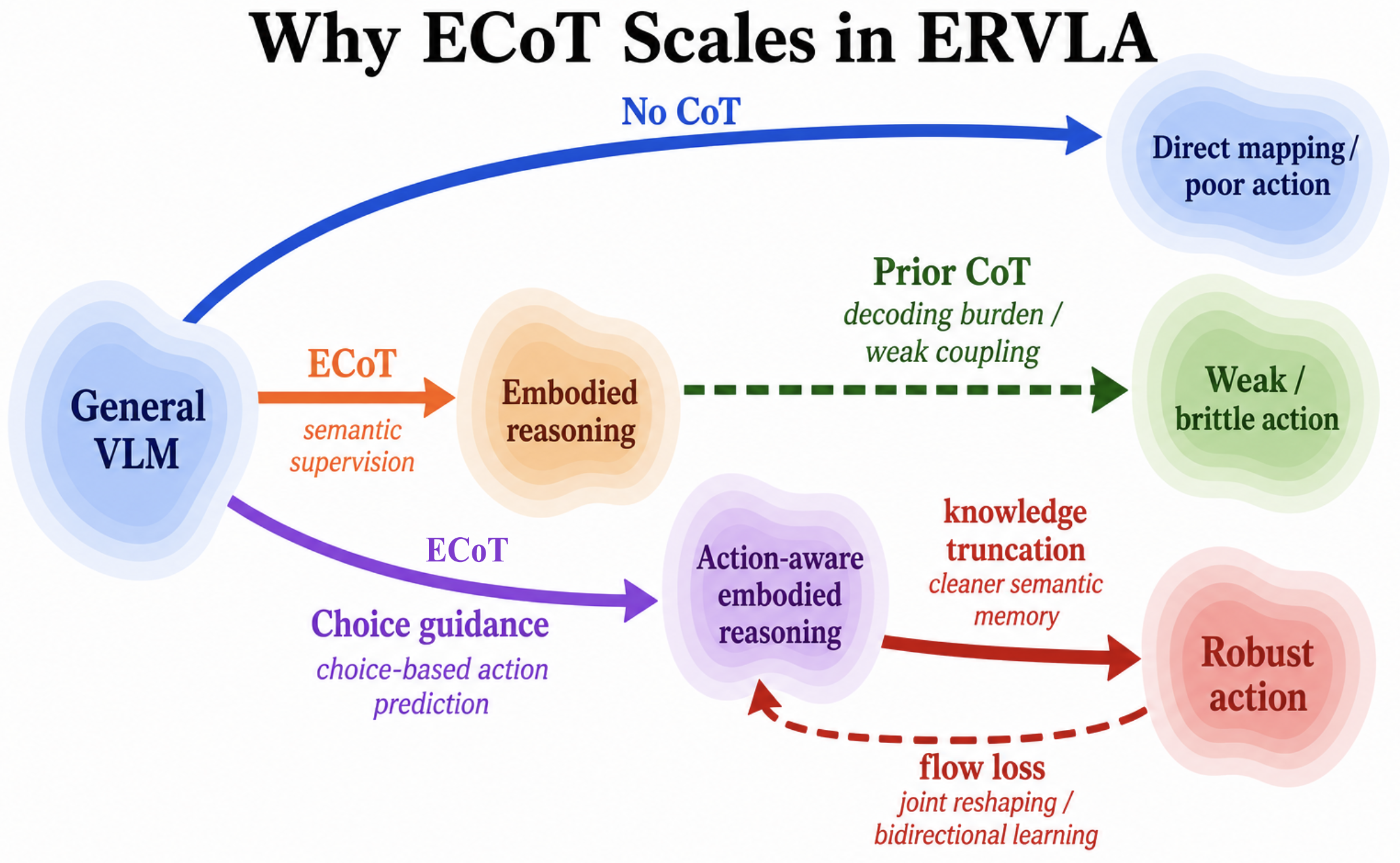}
\caption{Next-token action decoding after CoT is brittle, while knowledge insulation limits action feedback. Coupling ECoT with choice policy and flow supervision enables reasoning and action generation to co-adapt for robust control.}
    \label{fig:why}
    \vspace{-3mm}
\end{wrapfigure}

\begin{figure}[!t]
    \centering
    \includegraphics[width=1.0\textwidth]{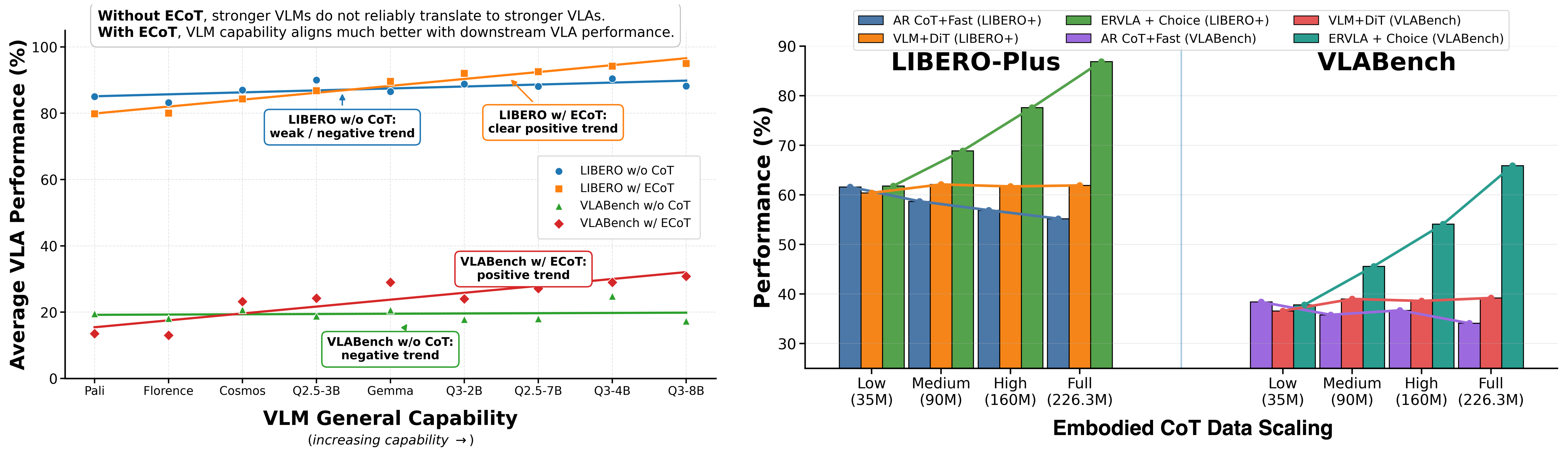}
    \caption{
    \textbf{VLM-to-VLA transfer and embodied CoT scaling.}
    \textbf{Left:} ECoT better aligns VLM capability with action.
    \textbf{Right:} ERVLA scales steadily with more CoT data on both LIBERO-Plus and VLABench, whereas AR CoT+Fast and isolated VLM+DiT show weaker or saturated scaling.
    }
    \label{fig:ecot_scaling_ablation}
\end{figure}

\noindent \textbf{ECoT scales through synergistic reasoning--action representation learning.}
The right panel of Figure.~\ref{fig:ecot_scaling_ablation} shows that ERVLA with the choice policy branch improves steadily as embodied CoT pre-training data increases on both LIBERO-Plus and VLABench. In contrast, autoregressive CoT+action-token designs, as in ECoT~\cite{zawalski2024robotic} and Emma-X~\cite{sun2024emmaxembodiedmultimodalaction}, become brittle because reasoning noise and length variability directly affect action decoding. An isolated VLM+DiT design, in the spirit of ThinkAct~\cite{huang2025thinkactvisionlanguageactionreasoningreinforced}, avoids this brittleness but weakly couples reasoning to action. Thus, the bottleneck is not whether embodied CoT is useful, but whether it can guide action learning without becoming a decoding burden. As depicted in Figure.~\ref{fig:why}, ERVLA uses CoT as high-level semantic supervision, choice policy prediction to internalize prefix reasoning into action-aware VLM representations, knowledge truncation to give DiT cleaner semantic memory, and flow loss for end-to-end action feedback. Embodied CoT therefore scales when reasoning supervision and action losses are jointly optimized in representation space, rather than treated as literal autoregressive targets.

\subsection{Real-world Experiments}
\label{sec:real-world}

To evaluate real-world manipulation beyond simulation, we deploy ERVLA on physical robots with one third-person and one wrist camera. We compare representative reasoning--action paradigms: ECoT~\cite{zawalski2024robotic} for autoregressive linguistic reasoning, WorldVLA~\cite{cen2025worldvlaautoregressiveactionworld} for using visual prediction as a reasoning--action interface, UniVLA~\cite{bu2025univlalearningacttaskcentric} for latent task-centric reasoning, and $\pi_{0.5}$~\cite{intelligence2025pi05visionlanguageactionmodelopenworld} as a strong VLM-conditioned DiT baseline with reasoning pre-training for continuous action generation.

\vspace{-2mm}

\paragraph{Task design.}
We design two task families for real-world experiments:
(i) placing items into drawers and
(ii) clearing tabletop waste.
The objects include toy cars, fruit models, bottles, and cans, together with a drawer and a basket.
We organize the evaluation into four difficulty tiers:

\begin{itemize}[leftmargin=*, itemsep=1pt, topsep=2pt]
    \item \textbf{Basic.}
    Clean scenes with explicit object names, mainly testing basic visual grounding and manipulation execution.

    \item \textbf{Distractors.}
    Irrelevant or visually similar objects are added to the scene, requiring the policy to identify the correct target while ignoring plausible distractors.

    \item \textbf{Semantic.}
    Instructions refer to objects indirectly through attributes, functions, categories, or spatial relations, requiring semantic interpretation beyond direct object-name matching.
    Examples include ``put away the object that is not a fruit,''
    ``place the blue car into the second drawer,''
    and ``remove the item that should be thrown away.''

    \item \textbf{Long-horizon.}
    Multiple challenges are combined into multi-object, multi-step tasks, requiring the policy to maintain task progress and execute a sequence of correct subgoals over an extended horizon.
\end{itemize}

Each tier contains five instructions, yielding 20 tasks in total.
We conduct five trials per task and report the mean success rate and progress score.

\begin{figure}[!t]
    \centering
    \includegraphics[width=1.0\textwidth]{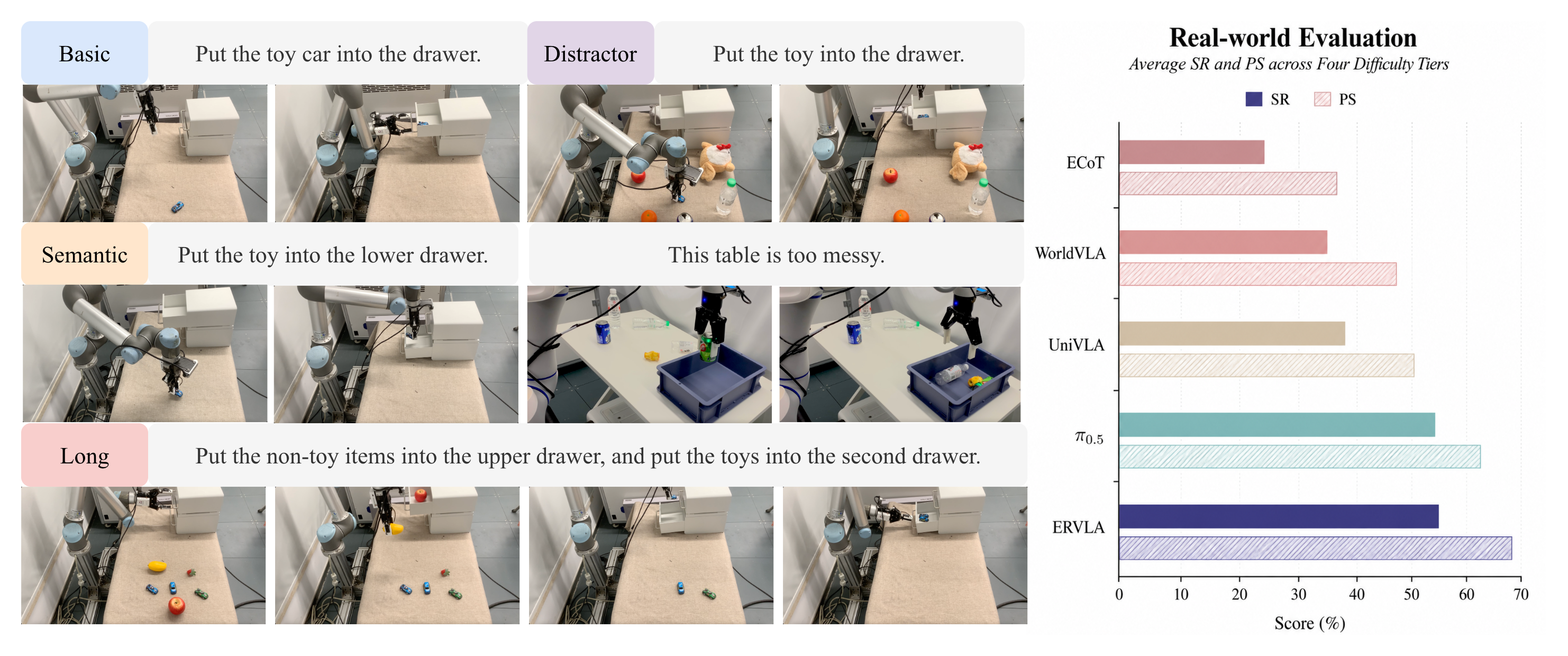}
    \caption{\textbf{Real-world evaluation.} Left: representative rollouts across basic, distractor, semantic, and long-horizon settings. Right: average success rate and progress score across four difficulty tiers, where ERVLA shows stronger robustness under semantic and long-horizon real-world generalization.}
    \label{fig:real_world_experiments}
\end{figure}

\begin{table}[t]
\centering
\small
\setlength{\tabcolsep}{2.4pt}
\renewcommand{\arraystretch}{1.08}
\caption{Real-world results across four difficulty tiers. Each tier reports success rate and progress score. Results are averaged over two task families.}
\label{tab:real_world}
\resizebox{\linewidth}{!}{
\begin{tabular}{lcccccccccc}
\toprule
\multirow{2}{*}{\textbf{Method}}
& \multicolumn{2}{c}{\textbf{Basic}}
& \multicolumn{2}{c}{\textbf{Distractors}}
& \multicolumn{2}{c}{\textbf{Semantic}}
& \multicolumn{2}{c}{\textbf{Long-horizon}}
& \multicolumn{2}{c}{\textbf{Average}} \\
\cmidrule(lr){2-3}
\cmidrule(lr){4-5}
\cmidrule(lr){6-7}
\cmidrule(lr){8-9}
\cmidrule(lr){10-11}
& \textbf{Succ. rate} & \textbf{Prog. score}
& \textbf{Succ. rate} & \textbf{Prog. score}
&  \textbf{Succ. rate} & \textbf{Prog. score}
& \textbf{Succ. rate} & \textbf{Prog. score}
& \textbf{Succ. rate} & \textbf{Prog. score} \\
\midrule
ECoT~\cite{zawalski2024robotic}
& 60 & 68
& 18 & 30
& 10 & 25
& 6  & 18
& 24 & 35 \\

WorldVLA~\cite{cen2025worldvlaautoregressiveactionworld}
& 78 & 84
& 28 & 42
& 18 & 35
& 12 & 28
& 34 & 47 \\

UniVLA~\cite{bu2025univlalearningacttaskcentric}
& 76 & 82
& 31 & 45
& 22 & 38
& 18 & 34
& 37 & 50 \\

$\pi_{0.5}$~\cite{intelligence2025pi05visionlanguageactionmodelopenworld}
& \textbf{97} & \textbf{98}
& \textbf{45} & 57
& 31 & 45
& 35 & 38
& 53 & 60 \\

ERVLA
& 96 & 97
& 44 & \textbf{58}
& \textbf{42} & \textbf{58}
& \textbf{38} & \textbf{55}
& \textbf{55} & \textbf{67} \\
\bottomrule
\end{tabular}}
\end{table}

\paragraph{Results.}
As depicted in Figure~\ref{fig:real_world_experiments}, ECoT~\cite{zawalski2024robotic} performs worst due to the latency and brittleness of autoregressive reasoning--action decoding, while WorldVLA~\cite{cen2025worldvlaautoregressiveactionworld} and UniVLA~\cite{bu2025univlalearningacttaskcentric} remain limited because visual prediction does not directly resolve semantic intent and latent reasoning can obscure fine-grained grounding. In contrast, ERVLA enables action-sensitive VLM representations and cleaner DiT conditioning. Detailed numbers in Table~\ref{tab:real_world} show that the main gap emerges not in clean cases, where $\pi_{0.5}$ ~\cite{intelligence2025pi05visionlanguageactionmodelopenworld}and ERVLA perform similarly, but under semantic ambiguity, distractors, and long-horizon dependencies. ERVLA gains most on the \textit{Semantic} and \textit{Long-horizon} tiers, supporting our claim that embodied CoT is most useful when internalized as action-relevant representation rather than emitted as mandatory explanation. Overall, these results indicate that real-world robustness comes less from stronger low-level action generation alone and more from whether reasoning supervision is converted into grounded, action-aware representations that remain useful under out-of-distribution semantic and temporal shifts.

\section{Conclusion}

In this work, we argue that embodied CoT should be understood less as an output format and more as an interface between semantic reasoning and action learning. The key lesson is that reasoning does not benefit robot policies simply by being longer, denser, or more explicit; it becomes useful only when its supervision is reliable, action-relevant, and integrated into the policy in a way that shapes representations rather than burdens decoding. From this perspective, both positive and negative of our findings are important: effective reasoning signals can expose the control value of pretrained VLM knowledge, while noisy grounding and naive autoregressive coupling reveal why reasoning-enhanced VLAs can fail to scale. ERVLA instantiates this principle by combining explicit linguistic CoT, reasoning dropout, VLM-level action shaping, and semantically conditioned continuous control into an end-to-end trainable recipe. More broadly, we hope that our dataset, annotation pipeline, model design, and systematic analyses provide a foundation for future work on reasoning-aware robot learning, including more reliable grounding supervision, adaptive test-time reasoning, stronger real-world generalization, and architectures that treat reasoning not as verbal decoration, but as a learnable bridge from perception to action.

\section*{Acknowledgments}

This work was conducted at Xiaomi Robotics with research and computing support from Xiaomi Robotics, and benefited from academic guidance and research support from Tsinghua University. The authors declare no competing interests.

\bibliographystyle{plainnat}
\bibliography{references}


\newpage
\appendix

\section{Related Work}
\label{appendix:related_work}

\subsection{Embodied Reasoning in Robot Manipulation}

Vision-language-action (VLA) models seek to turn pretrained visual-language knowledge into executable robot control. Early systems such as RT-2~\cite{zitkovich2023rt} and OpenVLA~\cite{kim2024openvla} demonstrate that web-scale semantic priors can be reused for manipulation, while recent generalist policies further scale this paradigm with larger backbones, broader embodiments, and stronger action heads, including $\pi_0$~\cite{black2026pi0visionlanguageactionflowmodel}, $\pi_{0.5}$~\cite{intelligence2025pi05visionlanguageactionmodelopenworld}, GR00T~\cite{bjorck2025gr00t}, Gemini Robotics~\cite{team2025gemini}, Eo-1~\cite{qu2025eo}, Hi Robot~\cite{shi2025hi}, and CogACT~\cite{li2024cogactfoundationalvisionlanguageactionmodel}. These models mark a transition from task-specific imitation policies to foundation-style robot policies, where language, vision, and prior world knowledge provide a broad semantic substrate for control. However, this transition also exposes a fundamental mismatch: VLMs are trained to produce symbolic or descriptive outputs, whereas robot policies must produce spatially precise, temporally consistent, and dynamically feasible actions.

The progress of VLAs is closely tied to large-scale robot data, including Open X-Embodiment~\cite{o2024open}, BridgeData V2~\cite{walke2024bridgedatav2datasetrobot}, DROID~\cite{khazatsky2025droidlargescaleinthewildrobot}, AgiBot World~\cite{agibotworldcontributors2025agibotworldcolosseolargescale}, MolmoAct~\cite{lee2025molmoactactionreasoningmodels}, and Fractal~\cite{zitkovich2023rt}. Yet scaling data and backbones alone does not fully resolve this semantic-to-action gap. VLM4VLA~\cite{zhang2026vlm4vla} shows that stronger VLMs do not necessarily yield stronger VLAs, suggesting that the bottleneck lies less in whether the model ``knows'' the task and more in whether such knowledge is transformed into action-relevant representations. In manipulation, useful knowledge must be grounded into objects, decomposed into subtasks, aligned with the robot state, and ultimately expressed as short-horizon motion. Embodied reasoning therefore serves as a missing interface: it converts high-level task understanding into structured signals that can guide perception-to-action learning, rather than relying on the policy to discover this bridge implicitly from raw trajectories alone.

\subsection{Taxonomy of Embodied Reasoning Signals}

To bridge high-level task understanding and motor execution, recent reasoning-aware VLAs introduce intermediate reasoning interfaces in linguistic, spatial, visual, and latent forms. A first line uses explicit linguistic or action-space CoT. ECoT~\cite{zawalski2024robotic} and EMMA-X~\cite{sun2024emmaxembodiedmultimodalaction} generate textual plans, subtasks, motion descriptions, and grounded coordinates before action prediction; LAP~\cite{zha2026laplanguageactionpretrainingenables} verbalizes low-level actions in natural language; ACoT-VLA~\cite{zhong2026acotvlaactionchainofthoughtvisionlanguageaction} reasons with coarse action intents; and ECoT-Lite studies efficient reasoning strategies such as reasoning dropout~\cite{chen2025trainingstrategiesefficientembodied}. Reinforced and verification-based methods, including Robot-R1~\cite{kim2025robot}, Embodied-R1~\cite{yuan2025embodied}, Mind-to-Hand~\cite{tang2025mindhandpurposefulrobotic}, and runtime reasoning-action alignment verification~\cite{wu2025you}, further show that reasoning becomes more useful when connected to execution feedback. These works make reasoning interpretable and directly supervisable, but also reveal that not all textual or grounded fields are equally reliable for control, and noisy explicit supervision can directly affect action learning.

A second line represents reasoning visually or through world-model-style prediction. CoT-VLA~\cite{zhao2025cotvlavisualchainofthoughtreasoning} predicts future visual subgoals, WorldVLA~\cite{cen2025worldvlaautoregressiveactionworld} uses visual prediction as a reasoning-action interface, Cosmos Policy~\cite{kim2026cosmos} adapts video models for visuomotor control, Mimic-Video~\cite{pai2025mimicvideovideoactionmodelsgeneralizable} studies video-action policies beyond standard VLAs, BridgeVLA~\cite{li2025bridgevlainputoutputalignmentefficient} uses 2D heatmap-style spatial supervision, and world action models investigate whether generative world models can act as zero-shot policies~\cite{ye2026world}. Related methods further explore spatial alignment and world-knowledge grounding~\cite{li2025spatialforcingimplicitspatial,zheng2026pokevlaempoweringpocketsizedvisionlanguageaction}. Visual reasoning naturally captures dynamics and spatial correspondence, but it is more expensive and harder to isolate: future images or heatmaps are slower to generate than text, often require per-step refresh, and may need inverse-dynamics or action-extraction modules to convert visual predictions into executable actions.

A third line compresses reasoning into latent or implicit representations. ThinkAct~\cite{huang2025thinkactvisionlanguageactionreasoningreinforced}, Fast-ThinkAct~\cite{huang2026fastthinkactefficientvisionlanguageactionreasoning}, LaST$_0$~\cite{liu2026last0latentspatiotemporalchainofthought}, and UniVLA~\cite{bu2025univlalearningacttaskcentric} distill reasoning into visual, verbalizable, spatio-temporal, or task-centric latents, while other systems move reasoning into policy representations rather than explicit traces~\cite{intelligence2025pi05visionlanguageactionmodelopenworld,li2024cogactfoundationalvisionlanguageactionmodel,shi2025hi,driess2025knowledgeinsulatingvisionlanguageactionmodels,wen2025dexvla}. These approaches reduce inference overhead, but usually assume that useful reasoning structure has already been learned or distilled. Therefore, our work begins from explicit embodied CoT not as the final deployment format, but as the most controllable interface for identifying which signals help action learning, which contaminate representations, and how reasoning should supervise continuous-action policies. This motivates using embodied CoT as representation-shaping pre-training rather than as a mandatory autoregressive prefix at inference time.

\subsection{Action Representations and Policy Integration}

A central design choice in VLA systems is how language-conditioned perception is converted into robot actions. Early models such as RT-2~\cite{zitkovich2023rt} and OpenVLA~\cite{kim2024openvla} formulate control as next-token prediction over discretized action tokens, reusing the VLM decoder to unify language and action generation. However, actions are not naturally linguistic objects: tokenization introduces quantization error, autoregressive decoding increases latency, and long reasoning prefixes make action prediction sensitive to exposure bias, reasoning noise, and sequence length. Methods such as OpenVLA-OFT~\cite{kim2025finetuningvisionlanguageactionmodelsoptimizing}, FAST~\cite{pertsch2025fastefficientactiontokenization}, VQ-VLA~\cite{wang2025vq}, BEAST~\cite{zhou2025beast}, and FASTer~\cite{liu2025faster} improve decoding or tokenization efficiency, but still keep reasoning and action tightly coupled through autoregressive token prediction.

Continuous-action policies instead model action chunks directly with imitation, diffusion, or flow objectives. ACT~\cite{zhao2023learning}, Diffusion Policy~\cite{chi2025diffusion}, RDT~\cite{liu2024rdt}, DexVLA~\cite{wen2025dexvla}, $\pi_0$~\cite{black2026pi0visionlanguageactionflowmodel}, and real-time action chunking flow policies~\cite{black2025realtimeexecutionactionchunking} better capture smoothness, multimodality, and temporal consistency. Yet this shifts the challenge from action representation to policy integration: the action expert must use VLM semantics without collapsing them into pooled-feature bottlenecks or damaging pretrained knowledge. Knowledge-insulated VLA training protects the backbone by blocking continuous-action gradients~\cite{driess2025knowledgeinsulatingvisionlanguageactionmodels}, but this weakens action feedback and limits the emergence of action-sensitive VLM representations. Other works improve the interface through spatial alignment, cross-embodiment prompting, or world knowledge, such as Spatial Forcing~\cite{li2025spatialforcingimplicitspatial}, X-VLA~\cite{zheng2025xvlasoftpromptedtransformerscalable}, and PokeVLA~\cite{zheng2026pokevlaempoweringpocketsizedvisionlanguageaction}.

Our work follows the continuous-action direction but focuses on the missing integration mechanism. Instead of treating CoT as an autoregressive action prefix, ERVLA uses embodied CoT as representation-shaping supervision. The choice branch injects action-level discrimination into the VLM, the DiT policy attends to per-layer VLM cache rather than compressed pooled features, and knowledge truncation restricts DiT conditioning to semantic-prefix memory rather than synthetic control-query tokens. This preserves end-to-end action feedback while avoiding shortcut conditioning. Thus, the key question is not only whether actions should be tokenized or generated continuously, but how reasoning supervision and action losses should jointly shape the representation that connects pretrained semantics to executable control.

\section{Embodied CoT Dataset Construction Pipeline}
\label{appendix:data_pipeline}

\begin{figure}[!t]
    \centering
    \includegraphics[width=1.0\textwidth]{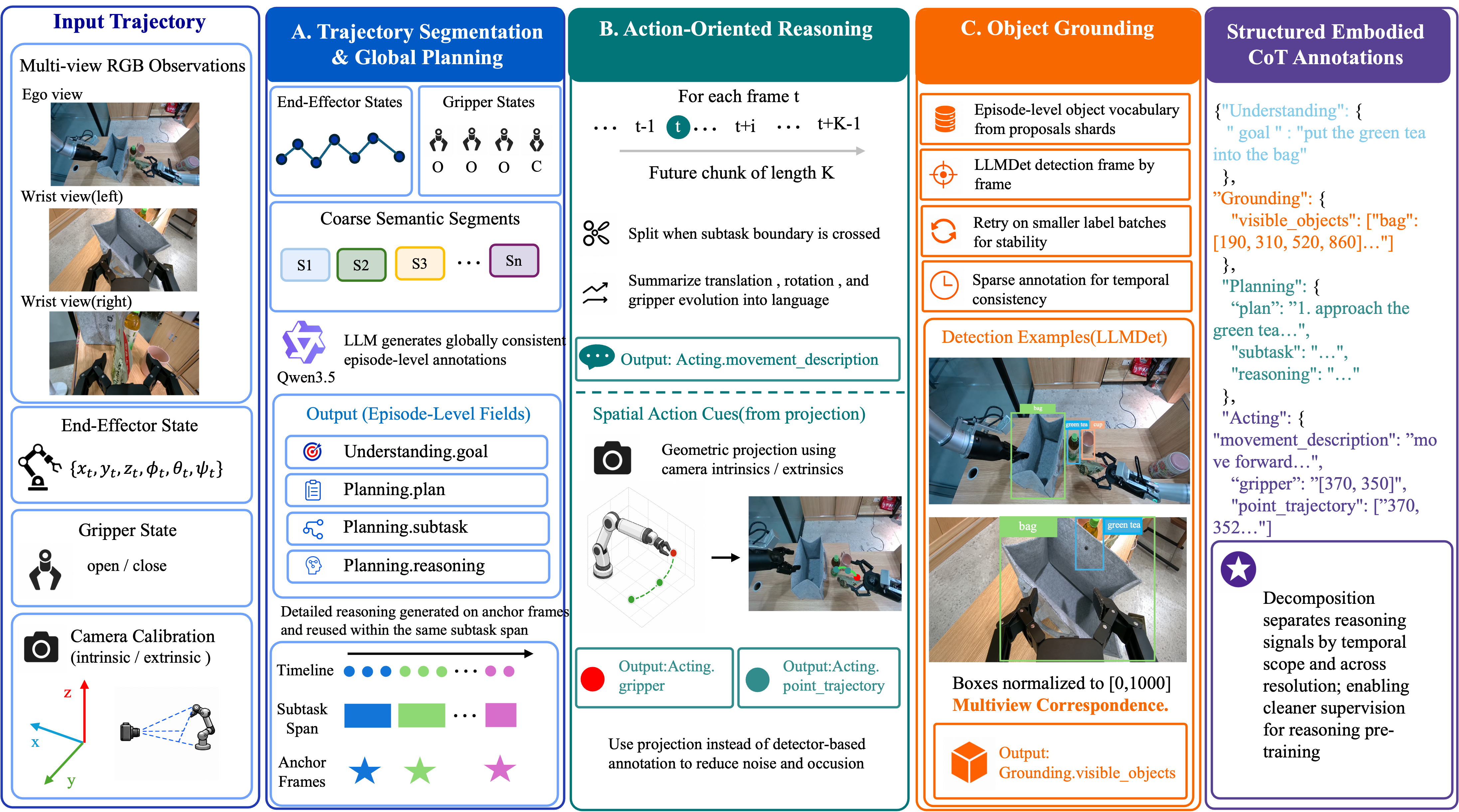}
    \caption{
    \textbf{Embodied CoT dataset construction pipeline.}
    Raw robot trajectories are converted into structured embodied reasoning supervision through trajectory segmentation, episode-level planning, action-oriented motion annotation, geometric gripper projection, future point-trajectory construction, and sparse object grounding. The resulting annotations provide hierarchical, grounded, and action-oriented CoT signals for scalable VLA pre-training.
    }
    \label{fig:data_pipeline}
\end{figure}

\paragraph{Overview.}
Given a raw robot trajectory, our goal is to convert low-level observations and actions into structured embodied CoT annotations that are useful for action learning. Each trajectory contains language instruction, multi-view RGB observations, proprioceptive states, gripper states, and low-level end-effector actions. We annotate each frame with four groups of reasoning signals: \texttt{Understanding}, \texttt{Grounding}, \texttt{Planning}, and \texttt{Acting}. The pipeline is staged rather than fully frame-independent: we first infer episode-level intent and subtask structure, then derive frame-level motion and grounding signals conditioned on this structure. This design reduces temporal inconsistency and avoids asking a language model to rediscover the same task logic at every frame. We demonstrate the entire annotation pipeline in Figure~\ref{fig:data_pipeline}.

\subsection{Trajectory Segmentation and Episode-Level Planning}
For each episode, we first partition the trajectory into coarse semantic segments using end-effector motion and gripper dynamics. Let the robot state at time $t$ be
\begin{equation}
\mathbf{s}_t = [\mathbf{p}_t, \mathbf{R}_t, g_t],
\end{equation}
where $\mathbf{p}_t \in \mathbb{R}^{3}$ is the end-effector position, $\mathbf{R}_t \in SO(3)$ is the end-effector orientation, and $g_t$ is the gripper state. We compute translational motion magnitude, rotational change, and gripper transitions:
\begin{equation}
\Delta p_t = \|\mathbf{p}_{t+1}-\mathbf{p}_t\|_2,\qquad
\Delta R_t = \|\log(\mathbf{R}_t^\top \mathbf{R}_{t+1})\|_2,\qquad
\Delta g_t = |g_{t+1}-g_t|.
\end{equation}
Candidate boundaries are created when the gripper opens or closes, when motion stops for a sustained window, or when the end-effector changes from approach-like motion to manipulation-like motion. These boundaries are used only as proposals, since low-level motion statistics alone cannot perfectly determine semantic subtask changes.

We then query Qwen3.5-397B~\cite{qwen35blog} once per episode using the task instruction, selected keyframes, and the proposed temporal segments. The VLM generates a globally consistent task goal and subtask sequence, including \texttt{Understanding.goal}, \texttt{Planning.plan}, \texttt{Planning.subtask}, and \texttt{Planning.reasoning}. This episode-level generation avoids frame-wise drift: all frames inherit the same global goal and plan, while the current subtask is assigned according to the segment that contains the frame. For long segments, we generate detailed subtask reasoning only on anchor frames and reuse it for neighboring non-anchor frames within the same subtask span. This reduces annotation cost while keeping the reasoning temporally coherent.

\subsection{Action-Oriented Reasoning From Future Motion}
For each frame, we construct action-oriented CoT from the future motion of the robot. Given a future horizon $K$, we extract the short-horizon action chunk
\begin{equation}
\mathbf{A}_{t:t+K} = [\mathbf{a}_t, \mathbf{a}_{t+1}, \ldots, \mathbf{a}_{t+K-1}],
\end{equation}
and truncate it if the chunk crosses a subtask boundary. We summarize the chunk into a compact language description stored as \texttt{Acting.movement\_description}. The description captures the dominant translation direction, rotation trend, and gripper evolution. For example, a chunk can be converted into a phrase such as ``move back 3 cm, move left 9 cm, move down 3 cm, keep gripper open.'' We use thresholded displacement and gripper-state changes to avoid over-describing small jitter:
\begin{equation}
\Delta \mathbf{p}_{t:t+K} = \mathbf{p}_{t+K}-\mathbf{p}_t,\qquad
\Delta g_{t:t+K}=g_{t+K}-g_t.
\end{equation}
Only motion components whose magnitude exceeds a small threshold are verbalized. This makes the movement field stable and action-relevant, rather than a noisy transcription of every small controller fluctuation.

\subsection{Geometric Gripper Projection}
For datasets with calibrated external cameras and end-effector poses, we annotate \texttt{Acting.gripper} through geometric projection rather than detector-based localization. This is important because grippers are often occluded, motion-blurred, or visually ambiguous; directly detecting them from images can introduce large spatial jitter. We apply this projection mainly to static third-person cameras, such as base or front views. For wrist-mounted cameras, the camera moves rigidly with the end-effector, making the current gripper location nearly fixed in the wrist-camera frame; therefore, we do not use wrist-camera gripper projection as action-oriented supervision.

For a static camera $c$, let its intrinsic matrix be
\begin{equation}
\mathbf{K}_c =
\begin{bmatrix}
f_x & 0 & c_x \\
0 & f_y & c_y \\
0 & 0 & 1
\end{bmatrix},
\end{equation}
and let $\mathbf{T}^{w}_{c} \in SE(3)$ denote the camera-to-world extrinsic matrix. The corresponding world-to-camera transform is
\begin{equation}
\mathbf{T}^{c}_{w} = (\mathbf{T}^{w}_{c})^{-1}.
\end{equation}

We define a gripper reference point in the end-effector frame, such as the tool center point or a calibrated fingertip midpoint, as
\begin{equation}
\bar{\mathbf{p}}^{ee}_{\mathrm{grip}}
=
[\mathbf{p}^{ee}_{\mathrm{grip}}, 1]^\top .
\end{equation}
Its world coordinate at time $t$ is
\begin{equation}
\bar{\mathbf{p}}^{w}_{\mathrm{grip}}(t)
=
\mathbf{T}^{w}_{ee}(t)
\bar{\mathbf{p}}^{ee}_{\mathrm{grip}},
\end{equation}
where $\mathbf{T}^{w}_{ee}(t)$ is the end-effector pose in the world frame. The point is then projected into camera $c$ by
\begin{equation}
\lambda
\begin{bmatrix}
u_t^c \\ v_t^c \\ 1
\end{bmatrix}
=
\mathbf{K}_c
\begin{bmatrix}
\mathbf{I}_{3 \times 3} & \mathbf{0}
\end{bmatrix}
\mathbf{T}^{c}_{w}
\bar{\mathbf{p}}^{w}_{\mathrm{grip}}(t).
\end{equation}
Equivalently, if the point in the camera coordinate frame is
$\mathbf{p}^{c}_{\mathrm{grip}}=(X,Y,Z)^\top$, then
\begin{equation}
u_t^c = f_x\frac{X}{Z}+c_x,\qquad
v_t^c = f_y\frac{Y}{Z}+c_y.
\end{equation}
A projection is considered valid only when $Z>0$ and the pixel lies inside the image. We normalize valid pixel coordinates to a $[0,1000]$ coordinate system:
\begin{equation}
\tilde{u}_t^c = \left\lfloor 1000 \frac{u_t^c}{W_c} \right\rceil,\qquad
\tilde{v}_t^c = \left\lfloor 1000 \frac{v_t^c}{H_c} \right\rceil,
\end{equation}
where $W_c$ and $H_c$ are the image width and height. The normalized coordinate $(\tilde{u}_t^c,\tilde{v}_t^c)$ is stored as \texttt{Acting.gripper} for valid external camera views. Invalid projections are omitted rather than replaced by unreliable detector outputs.

\subsection{Future Point-Trajectory Annotation}
We construct \texttt{Acting.point\_trajectory} by projecting future gripper positions into the image plane. For each frame $t$, we select a sequence of future timestamps $\{t+\delta_1,\ldots,t+\delta_M\}$ and project the gripper point at each timestamp using the same camera model:
\begin{equation}
\mathcal{P}^{c}_{t}
=
[
(\tilde{u}^{c}_{t+\delta_1}, \tilde{v}^{c}_{t+\delta_1}),
\ldots,
(\tilde{u}^{c}_{t+\delta_M}, \tilde{v}^{c}_{t+\delta_M})
].
\end{equation}
To make the trajectory stable, we skip consecutive duplicate or near-duplicate projected points, since they often correspond to stationary frames or controller jitter. If fewer than $M$ valid future points remain, we pad the sequence with $[-1,-1]$. This field provides the model with a spatially grounded motion trend, but avoids forcing dense per-frame pixel supervision when the projection is unreliable.

\subsection{Object Grounding}
For task-relevant object grounding, we use LLMDet~\cite{fu2025llmdetlearningstrongopenvocabulary} to produce bounding boxes. We first build a compact episode-level vocabulary of candidate objects from the instruction, generated plan, subtask descriptions, and detected proposal shards. This step prevents the detector from being queried with an overly large and unstable open vocabulary. For each frame and camera view, we query the detector with small batches of object names and retry failed or low-confidence detections with reduced label sets.

Given a detected box in image coordinates,
\begin{equation}
\mathbf{b} = [x_1,y_1,x_2,y_2],
\end{equation}
we normalize it to the same $[0,1000]$ coordinate system:
\begin{equation}
\tilde{\mathbf{b}}
=
\left[
\left\lfloor 1000\frac{x_1}{W}\right\rceil,
\left\lfloor 1000\frac{y_1}{H}\right\rceil,
\left\lfloor 1000\frac{x_2}{W}\right\rceil,
\left\lfloor 1000\frac{y_2}{H}\right\rceil
\right].
\end{equation}
The normalized boxes are stored in \texttt{Grounding.visible\_objects}. Since dense object boxes generated independently for every frame can fluctuate strongly across nearly identical observations, we adopt sparse grounding annotation. Specifically, object boxes are annotated on keyframes or at a fixed sparse interval, while frames without reliable detections omit the field. This choice is deliberate: for reasoning pre-training, missing uncertain grounding is less harmful than imposing inconsistent spatial labels on semantically similar frames.

\subsection{Multi-view Correspondence and Bimanual Formatting}
A distinctive feature of our dataset is that spatial CoT fields are not stored as a single-view annotation, but as view-indexed multi-view correspondence. To the best of our knowledge, this is the first embodied CoT dataset at this scale to explicitly preserve such multi-view spatial reasoning signals across base, front, and wrist cameras. Specifically, \texttt{Grounding.visible\_objects}, \texttt{Acting.gripper}, and \texttt{Acting.waypoint\_px} are organized by camera view, so that the same semantic object or future motion trend can be grounded differently under different viewpoints (see Figure~\ref{fig:multiview_cot_example}). This allows the model to learn view-consistent reasoning instead of overfitting to a single camera layout.

\begin{figure}[!t]
\centering
\begin{minipage}{0.48\textwidth}
\begin{lstlisting}[style=jsonstyle]
"Grounding": {
  "visible_objects": {
    "Base View": {
      "table": [0, 644, 1000, 1000],
      "fridge_open": [0, 0, 771, 696],
      "cola": [146, 321, 296, 602],
      "juice": [402, 208, 502, 604]
    },
    "Front View": {
      "table": [0, 577, 1000, 1000],
      "fridge_open": [258, 144, 888, 654],
      "cola": [625, 504, 673, 596],
      "juice": [692, 479, 727, 608]
    },
    "Wrist View": {
      "table": [0, 0, 1000, 558],
      "fridge_open": [985, 471, 1000, 519]
    }
  }
}
\end{lstlisting}
\end{minipage}
\hfill
\begin{minipage}{0.48\textwidth}
\begin{lstlisting}[style=jsonstyle]
"Acting": {
  "point_trajectory": {
    "Base View": [
      [459, 23], [458, 52],
      [457, 79], [456, 106],
      [455, 133], [455, 159],
      [454, 182], [453, 209],
      [452, 231], [452, 257]
    ],
    "Front View": [
      [388, 217], [396, 221],
      [407, 225], [418, 229],
      [429, 232], [436, 238],
      [442, 247], [448, 255],
      [454, 263], [460, 270]
    ]
  }
}
\end{lstlisting}
\end{minipage}
\vspace{-1mm}
\caption{\textbf{Example of multi-view spatial CoT annotation.} The same objects and future gripper motion are represented under different camera views, enabling view-consistent grounding and action-oriented reasoning. Point trajectory is removed from wrist views.}
\label{fig:multiview_cot_example}
\vspace{-2mm}
\end{figure}

For bimanual trajectories, we further preserve arm identity in action-oriented fields. End-effector states, gripper projections, and future waypoint trajectories are computed separately for the left and right arms when both are available. For single-arm data, the same schema is retained with only the active arm populated. This unified format allows single-arm, dual-arm, single-view, and multi-view data to share one language interface for reasoning pre-training, while preserving the spatial structure needed for grounded action learning.

\subsection{Quality Control and Filtering}
We apply several filtering rules before writing the final annotations. First, trajectories with missing images, invalid states, or inconsistent action lengths are removed. Second, projected gripper points are discarded if they fall outside the image, have negative depth, or jump unrealistically between adjacent frames. Third, object boxes with invalid geometry, extremely small area, or unstable category assignment are removed. Fourth, CoT fields are pruned if they are empty or inconsistent with the available observations. These filters are important because our experiments show that unreliable grounding can cause CoT contamination; therefore, the pipeline favors sparse but trustworthy supervision over dense but noisy labels.

The final output is a frame-level JSON-style annotation paired with the original trajectory data. A typical sample contains the task instruction, multi-view images, robot state, action chunk, and structured CoT fields:
\begin{equation}
\mathcal{D}_t =
\{
I_t^{1:C},\; \mathbf{s}_t,\; \mathbf{A}_{t:t+K},\;
\texttt{Understanding},\;
\texttt{Grounding},\;
\texttt{Planning},\;
\texttt{Acting}
\}.
\end{equation}
This format supports both explicit CoT supervision and reasoning dropout during training. It also allows different fields to be selectively ablated, enabling controlled analysis of which embodied reasoning signals are helpful, harmful, or scalable.

Overall, the pipeline separates reasoning signals by temporal scope and noise sensitivity. Episode-level fields provide stable semantic intent and task structure; frame-level movement fields connect reasoning to short-horizon control; geometric projection provides cleaner gripper supervision when calibration is available; and sparse object grounding avoids injecting dense detector noise. Based on this pipeline, we construct a large-scale embodied CoT corpus comprising \textbf{978,743} trajectories, \textbf{226.3M} samples, and \textbf{2592.5} hours of data across Bridge~\cite{walke2024bridgedatav2datasetrobot}, Fractal~\cite{zitkovich2023rt}, Droid~\cite{khazatsky2025droidlargescaleinthewildrobot}, MolmoAct~\cite{lee2025molmoactactionreasoningmodels}, and AgiBot~\cite{agibotworldcontributors2025agibotworldcolosseolargescale} as depicted in Figure.~\ref{fig:dataset}. To the best of our knowledge, it is the largest embodied CoT dataset to date and one of the first at this scale to include both multi-view and bimanual settings.

\section{Training Details}
\label{appendix:training_details}

This section provides implementation and training details for all experiments in the paper.
We organize the discussion into three groups: (i) controlled embodied CoT studies with autoregressive action-token prediction, (ii) VLM-to-VLA transfer studies, and (iii) ERVLA pre-training and ablations. 

\subsection{Pre-training Data}
\label{appendix:pretraining_data}

\begin{table}[!t]
\centering
\small
\setlength{\tabcolsep}{5.5pt}
\renewcommand{\arraystretch}{1.08}
\caption{\textbf{Summary of pre-training robot CoT data.} Statistics are reported after trajectory cleaning, CoT matching, and quality filtering. The total excludes auxiliary VLM-only data, which is mixed only to preserve general multimodal capability.}
\label{tab:pretraining_data}
\resizebox{\linewidth}{!}{
\begin{tabular}{lccrrrl}
\toprule
\textbf{Dataset} 
& \textbf{Source} 
& \textbf{Action Horizon} 
& \textbf{Trajectories} 
& \textbf{Samples} 
& \textbf{Duration (h)} 
& \textbf{Notes} \\
\midrule

AgiBot World~\cite{agibotworldcontributors2025agibotworldcolosseolargescale}
& Real 
& 30 
& 765,649 
& 204.94M 
& 1879.4 
& Multi-view, bimanual manipulation \\

DROID~\cite{khazatsky2025droidlargescaleinthewildrobot}
& Real 
& 15 
& 74,682 
& 15.12M 
& 280.0 
& In-the-wild single-arm manipulation \\

Fractal~\cite{gaydon2024fractalultralargescaleaeriallidar}
& Real 
& 3 
& 86,703 
& 3.70M 
& 342.5 
& Large-scale tabletop manipulation \\

BridgeData V2~\cite{walke2024bridgedatav2datasetrobot}
& Real 
& 5 
& 43,807 
& 1.43M 
& 79.7 
& Tabletop manipulation and object interaction \\

MolmoAct~\cite{lee2025molmoactactionreasoningmodels}
& Real 
& 15
& 7,902 
& 1.10M 
& 10.9 
& single-arm Franka manipulation \\

\midrule
\textbf{Robot CoT total} 
& -- 
& -- 
& \textbf{978,743} 
& \textbf{226.3M} 
& \textbf{2592.5} 
& Excludes auxiliary VLM-only data \\

\midrule
Auxiliary VLM data 
& Web/VLM 
& -- 
& -- 
& -- 
& -- 
& Mixed at a small ratio\\

\bottomrule
\end{tabular}}
\end{table}

Table~\ref{tab:pretraining_data} summarizes the robot CoT data used for ERVLA pre-training. 
We clean and standardize large-scale robot datasets, then annotate them with structured embodied CoT fields, including task understanding, object grounding, planning, motion descriptions, and action-oriented spatial cues when available. 
The pre-training mixture contains both single-arm and bimanual manipulation, as well as single-view and multi-view observations. 
We also mix a small amount of general VLM instruction data to preserve the backbone's multimodal instruction-following ability; these auxiliary VLM samples do not contain robot trajectories and are not counted in the robot CoT totals.

\subsection{Post-training Data}
\label{appendix:posttraining_data}

In addition to large-scale robot CoT pre-training, we also annotate all downstream post-training data with the same structured ECoT schema. This ensures that benchmark adaptation does not rely on a different prompting or supervision format from pre-training. Specifically, LIBERO and VLABench post-training samples are converted into frame-level annotations containing \texttt{Understanding}, \texttt{Grounding}, \texttt{Planning}, and \texttt{Acting} fields. Each sample includes the task goal, visible object boxes, global plan, current subtask, subtask reasoning, movement description, gripper pose, and future waypoint trajectory when available. For multi-view settings, spatial fields are stored as view-indexed dictionaries so that the same object or motion cue can be grounded under different cameras.

Table~\ref{tab:posttraining_data} summarizes the post-training data used in our experiments. VLABench provides long-horizon and semantically diverse manipulation tasks, while LIBERO provides four post-training suites covering general, goal-conditioned, object-conditioned, and spatial transfer settings. All suites are annotated with the complete ECoT field set and use the same field names as the pre-training corpus. This unified format allows the model to reuse the same reasoning interface during pre-training, post-training, ablation, and evaluation.

\begin{table}[!t]
\centering
\small
\setlength{\tabcolsep}{6pt}
\renewcommand{\arraystretch}{1.08}
\caption{\textbf{Summary of post-training ECoT data.} All datasets are annotated with the same structured ECoT schema, including task understanding, object grounding, planning, subtask reasoning, movement description, gripper pose, and future waypoint trajectory.}
\label{tab:posttraining_data}
\resizebox{0.55\linewidth}{!}{
\begin{tabular}{lrr}
\toprule
\textbf{Dataset} 
& \textbf{Trajectories} 
& \textbf{Samples} \\
\midrule
VLABench~\cite{zhang2024vlabenchlargescalebenchmarklanguageconditioned}
& 5,000
& 575,101 \\

LIBERO-10~\cite{liu2023liberobenchmarkingknowledgetransfer}
& 500
& 138,090 \\

LIBERO-Goal~\cite{liu2023liberobenchmarkingknowledgetransfer}
& 500
& 63,728 \\

LIBERO-Object~\cite{liu2023liberobenchmarkingknowledgetransfer}
& 500
& 74,507 \\

LIBERO-Spatial~\cite{liu2023liberobenchmarkingknowledgetransfer}
& 500
& 62,250 \\

\midrule
\textbf{Total}
& \textbf{7,000}
& \textbf{913,676} \\
\bottomrule
\end{tabular}}
\end{table}

Figure.~\ref{fig:posttrain_ecot_example} shows one frame-level ECoT annotation from LIBERO-10. The example illustrates three important properties of our post-training format. First, object grounding is multi-view: the same objects are represented separately under \texttt{cam0} and \texttt{cam1}. Second, action-oriented fields provide both language-level motion description and spatial waypoint supervision. Third, planning fields are episode-consistent: the global plan is shared across the trajectory, while the current subtask and reasoning are attached to each frame through the corresponding subtask span. Non-anchor frames can reuse cached reasoning from an anchor frame, reducing redundant annotation while preserving temporal consistency.

\begin{figure}[!t]
\centering
\begin{minipage}{0.98\linewidth}
\begin{lstlisting}[style=jsonstyle]
{
  "Understanding": {
    "goal": "turn on the stove and put the moka pot on it"
  },
  "Grounding": {
    "visible_objects": {
      "cam0": {
        "chefmate_8_frypan_1": [457, 480, 1000, 754],
        "moka_pot_1": [395, 527, 688, 789],
        "flat_stove_1": [94, 418, 395, 754]
      },
      "cam1": {
        "chefmate_8_frypan_1": [4, 82, 570, 523],
        "moka_pot_1": [219, 0, 699, 219],
        "flat_stove_1": [672, 102, 1000, 707]
      }
    }
  },
  "Planning": {
    "plan": [
      {"id": 1, "subtask": "approach stove knob"},
      {"id": 2, "subtask": "grasp stove knob"},
      {"id": 3, "subtask": "turn stove knob"},
      {"id": 4, "subtask": "release stove knob and retreat"},
      {"id": 5, "subtask": "grasp moka pot"},
      {"id": 6, "subtask": "place moka pot on stove and retreat"}
    ],
    "subtask_name": "approach stove knob",
    "subtask_reasoning": "The robot is positioned above the counter with the stove knob, moka pot, and pan visible below. No actions have been completed yet as the gripper remains open. The robot needs to lower its arm to approach the stove knob for interaction."
  },
  "Acting": {
    "movement_description": "move left 4 cm, keep gripper open",
    "gripper_pose": {
      "cam0": [508, 234]
    },
    "waypoint_px": {
      "cam0": [
        [504, 234], [500, 234], [496, 234], [492, 234], [484, 234],
        [480, 234], [473, 234], [469, 234], [461, 234], [453, 234]
      ]
    },
    "subtask_end": 61
  }
}
\end{lstlisting}
\end{minipage}
\vspace{-1mm}
\caption{\textbf{Example frame-level ECoT annotation from LIBERO-10.} The annotation contains multi-view object grounding, episode-level planning, frame-level subtask reasoning, and action-oriented spatial cues. Coordinates are normalized to the $[0,1000]$ image coordinate system.}
\label{fig:posttrain_ecot_example}
\vspace{-2mm}
\end{figure}

\subsection{Controlled CoT Field Studies}
\label{appendix:controlled_cot_fields}

The first set of experiments studies which embodied CoT fields are useful for action learning. To isolate the effect of reasoning supervision, we use Qwen3-VL-4B~\cite{bai2025qwen3vltechnicalreport} as the unified backbone and adopt the same autoregressive CoT+FAST action-token interface~\cite{pertsch2025fastefficientactiontokenization} for all variants. Each sample is formatted as a multimodal instruction-following sequence: multi-view observations and the task instruction are provided as input, the model first predicts the selected CoT fields wrapped by \texttt{<cot>} and \texttt{</cot>}, and then predicts FAST action tokens through next-token prediction. The action horizon, tokenizer, data split, optimizer, and evaluation protocol are kept fixed, so that performance differences mainly reflect the contribution of different CoT fields.

We consider two training regimes. In the direct post-training regime, the model is initialized from Qwen3-VL-4B and trained only on the downstream VLABench data with the selected CoT fields. This setting tests whether a field provides immediate supervision for action learning without additional robot pre-training. In the Bridge pre-training regime, we first pre-train the model on BridgeData V2~\cite{walke2024bridgedatav2datasetrobot} annotated with the same structured CoT format, and then post-train it on VLABench. This setting tests whether each CoT field remains beneficial after the model has already learned a basic mapping between embodied reasoning and action tokens. For both regimes, we report changes relative to the corresponding no-CoT baseline.

We also evaluate reasoning dropout. During training, each sample uses explicit CoT with probability $p=0.5$; otherwise, the CoT span is replaced by an empty \texttt{<cot></cot>} block while the model is still supervised to predict the same action tokens. This prevents the policy from relying on explicit reasoning text as a mandatory action prefix and encourages reasoning information to be internalized into action-relevant hidden states. Comparing training with and without dropout allows us to separate useful reasoning supervision from brittle dependence on explicit CoT. These controlled studies show that low-level action-centric fields, such as movement descriptions and end-effector trajectories in 2D image space, are more directly useful for control than high-level task descriptions alone, while dense but noisy grounding fields can cause CoT contamination if used without dropout or field selection.

\subsection{VLM-to-VLA Transfer Study}
\label{appendix:vlm_to_vla_transfer}

The second set of experiments studies whether stronger VLMs become stronger VLAs when embodied CoT is used as a transfer interface. Following the controlled comparison principle of VLM4VLA, we keep the VLA adaptation recipe fixed across all backbones: the same StarVLA~\cite{community2026starvlalegolikecodebasevisionlanguageaction} training pipeline, FAST~\cite{pertsch2025fastefficientactiontokenization} action-token interface, action horizon, data preprocessing, optimizer family, augmentation pipeline, and evaluation tasks are used for every model. Each backbone is trained under two settings: without CoT supervision and with full embodied CoT supervision under reasoning dropout. This design isolates whether embodied CoT improves the conversion of pretrained semantic priors into action-relevant representations, rather than confounding the comparison with different policy heads, action tokenizers, or data processing strategies.

For each VLM, we preserve its native multimodal instruction format as much as possible. Images are inserted using the model-specific vision-token convention, while the robot instruction and optional embodied CoT fields are formatted as the text context. For CoT-enabled training, the assistant target first contains the structured embodied CoT block and then the FAST action tokens; for the no-CoT setting, the CoT span is replaced by an empty \texttt{<cot></cot>} block while the same action tokens are still supervised. Reasoning dropout is applied with probability $p=0.5$, so the model sees both explicit-CoT and no-CoT inputs during training. This prevents the policy from treating CoT as a mandatory autoregressive prefix and instead encourages the VLM hidden states to internalize reasoning information useful for action prediction.

All VLM-to-VLA transfer runs fine-tune the VLM backbone together with the FAST action-token prediction head, rather than freezing the vision encoder or language model. Following the controlled comparison principle of VLM4VLA, we use the same StarVLA adaptation recipe across all backbones. To keep the comparison fair, we fix the action-token interface, data split, optimizer, augmentation, checkpointing, and evaluation protocol (see Table~\ref{tab:vlm_to_vla_training_protocol}). For this transfer study, we use short benchmark-level training schedules: 10,000 steps on LIBERO and 20,000 steps on VLABench.

\begin{table}[!t]
\centering
\small
\setlength{\tabcolsep}{5.5pt}
\renewcommand{\arraystretch}{1.08}
\caption{\textbf{Training protocol for the VLM-to-VLA transfer study.} All backbones use the same StarVLA adaptation recipe; only the VLM backbone and CoT setting are changed.}
\label{tab:vlm_to_vla_training_protocol}
\resizebox{\linewidth}{!}{
\begin{tabular}{ll}
\toprule
\textbf{Item} & \textbf{Setting} \\
\midrule
Framework & StarVLA unified training pipeline \\
Action interface & FAST action-token prediction \\
Backbones & PaliGemma-2, Florence-2, Cosmos-Reason2, Gemma, Qwen2.5-VL, Qwen3-VL \\
CoT settings & w/o CoT; w/ full ECoT + reasoning dropout \\
Reasoning dropout & $p=0.5$ \\
Trainable modules & Vision encoder, LLM, token embeddings, and action-token head \\
Observation & Current visual observation with task instruction \\
Image preprocessing & Resize with \texttt{max\_pixels}=90k, random crop ratio 0.95, photometric distortion \\
Action horizon & 10 \\
Action representation & FAST tokens from 7-DoF end-effector action chunks \\
Action-token loss & Next-token prediction on FAST action tokens \\
Optimizer & AdamW \\
Adam betas & $(0.9, 0.95)$ \\
Weight decay & 0.1 \\
Peak learning rate & $1\times10^{-4}$ \\
Minimum learning rate & $5\times10^{-7}$ \\
Warmup steps & 2,000 \\
LR schedule & Cosine decay \\
Gradient clipping & 1.0 \\
Precision & bfloat16 mixed precision \\
LIBERO training steps & 10,000 \\
VLABench training steps & 20,000 \\
Checkpoint interval & 2,000 steps \\
\bottomrule
\end{tabular}}
\vspace{-2mm}
\end{table}

We evaluate a diverse set of VLM backbones spanning different model families and capability levels, including PaliGemma-2-3B~\cite{beyer2024paligemmaversatile3bvlm}, Florence-2-large~\cite{xiao2023florence2advancingunifiedrepresentation}, Cosmos-Reason2-2B~\cite{nvidia2025cosmosreason1physicalcommonsense}, Gemma-4-E2B-it, Qwen2.5-VL-3B/7B~\cite{qwen2025qwen25technicalreport}, and Qwen3-VL-2B/4B/8B~\cite{bai2025qwen3vltechnicalreport}. Table~\ref{tab:vlm_backbone_capability_order} orders these backbones by general VLM capability. Since all models share the same action-token decoder and training recipe, this study directly tests whether embodied CoT makes pretrained VLM capability more transferable to VLA control.

\begin{table*}[!t]
\centering
\scriptsize
\setlength{\tabcolsep}{4.0pt}
\renewcommand{\arraystretch}{1.08}
\caption{
\textbf{VLM backbone capability order used in the VLM-to-VLA transfer study.}
Models are roughly ordered from weaker to stronger general VLM capability based on available public VLM evaluation results.
Scores are reported when comparable results are available.
}
\label{tab:vlm_backbone_capability_order}
\resizebox{\textwidth}{!}{
\begin{tabular}{lccccccccc}
\toprule
\textbf{Backbone}
& \textbf{Param.}
& \textbf{Avg.}
& \textbf{MMBench}
& \textbf{MMStar}
& \textbf{MMMU}
& \textbf{MathVista}
& \textbf{OCRBench}
& \textbf{AI2D}
& \textbf{HallusionBench} \\
\midrule

PaliGemma-2-3B~\cite{beyer2024paligemmaversatile3bvlm}
& 3B
& 46.5
& 65.6
& 48.3
& 34.9
& 28.5
& 614
& 68.3
& 32.2 \\

Florence-2-large~\cite{xiao2023florence2advancingunifiedrepresentation}
& 0.77B
& 52.0
& 69.5
& --
& 39.0
& --
& 690
& --
& 35.0 \\

Cosmos-Reason2-2B~\cite{nvidia2025cosmosreason1physicalcommonsense}
& 2B
& 61.8
& 74.0
& --
& 48.5
& 57.0
& --
& 76.0
& -- \\

Gemma-4-E2B-it
& 2B
& 63.2
& 75.5
& 56.8
& --
& 59.5
& --
& 78.0
& -- \\

Qwen2.5-VL-3B~\cite{qwen2025qwen25technicalreport}
& 3B
& 64.5
& 76.8
& 56.3
& 51.2
& 61.2
& 828
& 81.4
& 46.6 \\

Qwen3-VL-2B~\cite{bai2025qwen3vltechnicalreport}
& 2B
& 66.5
& 78.4
& 58.3
& 53.4
& 61.3
& 858
& 76.9
& 51.4 \\

Qwen2.5-VL-7B~\cite{qwen2025qwen25technicalreport}
& 7B
& 70.9
& 82.2
& 64.1
& 58.0
& 68.1
& 888
& 84.3
& 51.9 \\

Qwen3-VL-4B~\cite{bai2025qwen3vltechnicalreport}
& 4B
& 74.94
& 83.9
& 69.8
& 67.4
& 73.7
& 881
& 84.1
& 57.6 \\

Qwen3-VL-8B~\cite{bai2025qwen3vltechnicalreport}
& 8B
& \textbf{76.94}
& \textbf{84.5}
& \textbf{70.9}
& \textbf{69.6}
& \textbf{77.2}
& \textbf{896}
& \textbf{85.7}
& \textbf{61.1} \\

\bottomrule
\end{tabular}}
\vspace{-2mm}
\end{table*}

\begin{table}[!t]
\centering
\scriptsize
\setlength{\tabcolsep}{4.5pt}
\renewcommand{\arraystretch}{1.02}
\caption{\textbf{Large-scale ERVLA pre-training hyperparameters.}}
\label{tab:appendix_pretrain_hyperparams}
\resizebox{0.72\linewidth}{!}{
\begin{tabular}{lc}
\toprule
\textbf{Item} & \textbf{Value} \\
\midrule
Backbone & Qwen3-VL-4B-Instruct \\
Action horizon & 30 \\
Action / state shape & $30 \times 60$ / $1 \times 60$ \\
Number of choices & 5 \\
DiT layers & 36 \\
Reasoning dropout & $p_{\mathrm{cot}}=0.5$ \\
CoT fields & Understanding, Grounding, Planning, Acting \\
Obs. length / interval & 1 / 1 \\
Image augmentation & Resize, random crop, photometric distortion \\
Max image pixels & 90,000 \\
Token packing & Enabled \\
Max packed length & 17,600 tokens \\
Training steps & 120,000 \\
Batch size & 64 \\
Precision & bfloat16 mixed precision \\
Optimizer & DeepSpeed FusedAdam / AdamW-style \\
Adam betas & $(0.9, 0.95)$ \\
Weight decay & 0.1 \\
Learning rate & $5\times10^{-5}$ \\
Warmup steps & 2,000 \\
Gradient clipping & 1.0 \\
Distributed strategy & DeepSpeed \\
Checkpoint / validation interval & 20,000 / 10,000 steps \\
\bottomrule
\end{tabular}}
\vspace{-2mm}
\end{table}

\subsection{Large-scale ERVLA Pre-training}

ERVLA is trained in two stages. First, we perform large-scale embodied CoT pre-training on a mixture of open robot datasets. ERVLA uses Qwen3-VL-4B as the reasoning backbone, a choice branch for candidate continuous action chunks, and a DiT policy for continuous action generation. The final model is trained end-to-end: diffusion loss, choice loss, score loss, and CoT token loss are optimized jointly unless an ablation explicitly blocks gradients.

The model receives semantic inputs consisting of images, instruction, and optional CoT, followed by trainable control-query tokens. The VLM produces hidden states and key-value caches. The choice branch reads action-query hidden states to predict $N$ candidate action chunks and their scores, while DiT generates the final continuous action chunk by attending to the knowledge-truncated semantic-prefix cache. In all ERVLA experiments, we use $N=5$ choices and a 36-layer DiT; full pre-training hyperparameters are provided in Table~\ref{tab:appendix_pretrain_hyperparams}.

For large-scale reasoning pre-training, we train on a mixture of embodied CoT data constructed from Bridge~\cite{walke2024bridgedatav2datasetrobot}, Fractal~\cite{zitkovich2023rt}, Droid~\cite{khazatsky2025droidlargescaleinthewildrobot}, MolmoAct~\cite{lee2025molmoactactionreasoningmodels}, and AgiBot~\cite{agibotworldcontributors2025agibotworldcolosseolargescale}. Compared with the base no-CoT pre-training configuration, we enable CoT loading with the same structured fields used in the main paper and apply reasoning dropout with $p_{\mathrm{cot}}=0.5$.

For single-arm datasets, actions are composed into a padded action vector. For dual-arm datasets such as AgiBot, we compose left-arm and right-arm end-effector position, axis-angle rotation, and gripper signals into a unified action vector. In the full pre-training configuration, actions are padded to 60 dimensions and states are padded to 60 dimensions, allowing a shared action interface across single-arm and dual-arm data. We use streaming MDS datasets and token packing to improve training efficiency. Packing allows multiple shorter multimodal samples to be concatenated into a long sequence up to the maximum training length, reducing padding waste and increasing effective token throughput.

As shown in Table~\ref{tab:appendix_pretrain_data_mixture}, we use a nominal sampling mixture to balance large-scale coverage and dataset diversity. AgiBot receives the largest sampling weight due to its scale, while smaller auxiliary datasets such as MolmoAct are sampled at lower rates to avoid over-representation.

\begin{table}[!t]
\centering
\scriptsize
\setlength{\tabcolsep}{5pt}
\renewcommand{\arraystretch}{1.02}
\caption{\textbf{Pre-training data mixture.} We report nominal sampling weights used to mix different embodied CoT sources during pre-training.}
\label{tab:appendix_pretrain_data_mixture}
\resizebox{0.62\linewidth}{!}{
\begin{tabular}{lrrr}
\toprule
\textbf{Dataset} & \textbf{Traj.} & \textbf{Samples} & \textbf{Weight} \\
\midrule
AgiBot~\cite{agibotworldcontributors2025agibotworldcolosseolargescale} & 765.6K & 204.9M & 0.518 \\
DROID~\cite{khazatsky2025droidlargescaleinthewildrobot} & 74.7K & 15.1M & 0.180 \\
Fractal~\cite{zitkovich2023rt} & 86.7K & 3.7M & 0.120 \\
Bridge~\cite{walke2024bridgedatav2datasetrobot} & 43.8K & 1.4M & 0.100 \\
MolmoAct~\cite{lee2025molmoactactionreasoningmodels} & 7.9K & 1.1M & 0.082 \\
\midrule
\textbf{Total} & \textbf{978.7K} & \textbf{226.3M} & \textbf{1.00} \\
\bottomrule
\end{tabular}}
\vspace{-2mm}
\end{table}

\subsection{Post-training}

For LIBERO post-training, we train on LIBERO-10, LIBERO-Spatial, LIBERO-Object, and LIBERO-Goal.
We use delta action targets from the dataset and compose them into the same 60-dimensional padded action vector used in ERVLA pre-training.
The action horizon is 10.
The robot state contains the end-effector and gripper state, padded to 60 dimensions.
We use Gaussian normalization for actions and states.
The same CoT fields are enabled with $p_{\mathrm{cot}}=0.5$.

The LIBERO post-training configuration is used both for the final LIBERO-Plus transfer evaluation and for ablations. All models are post-trained only on LIBERO and evaluated zero-shot on LIBERO-Plus. This setting tests whether reasoning pre-training improves robustness beyond the original LIBERO distribution.

For VLABench post-training, we train on the VLABench CoT dataset.
The action horizon is 10.
Unlike LIBERO, VLABench uses delta end-effector actions derived from absolute end-effector states and actions.
We wrap Euler angles when computing rotational deltas and compose each action into the same 60-dimensional padded action space.
We use Gaussian normalization for both actions and proprioceptive states, which are also padded to 60 dimensions.
The \texttt{use\_ideal\_length} option is enabled so that training samples satisfy the required action horizon for choice and diffusion supervision.

\subsection{ERVLA Ablations}
We use the same pre-training and post-training data for ERVLA ablations unless otherwise stated. The ablations are designed to isolate the role of the choice branch, end-to-end gradient flow, and knowledge truncation. 

\paragraph{No Choice (End-to-End).}
This variant removes the auxiliary choice branch and trains the VLM+DiT model end-to-end. Diffusion loss is still allowed to update the VLM backbone. This tests whether end-to-end continuous-action training alone is sufficient without candidate-level action discrimination.

\paragraph{No Choice + Knowledge Insulation.}
This variant removes the choice branch and blocks gradients from the continuous action expert to the VLM backbone. It resembles planning-action separation in which the VLM representation is preserved but receives weaker feedback from action generation. This tests whether preserving pretrained knowledge by stopping gradients is preferable to end-to-end action shaping.

\paragraph{Choice + No Knowledge Truncation.}
This variant keeps the choice branch but removes knowledge truncation. DiT is allowed to condition on the full VLM cache, including appended state and action-query turns. This tests whether DiT exploits synthetic control-query tokens as shortcuts when the full cache is exposed.

\paragraph{Full ERVLA.}
The full model combines explicit CoT supervision, reasoning dropout, the choice policy branch, knowledge truncation, and end-to-end diffusion training. DiT conditions only on the semantic-prefix cache, while gradients from action generation still update the VLM backbone. This allows CoT learning, choice-based action shaping, and continuous action generation to co-adapt.

\subsection{Practical Notes}

All training runs use bfloat16 mixed precision and DeepSpeed distributed training. We use gradient clipping with a maximum norm of 1.0 and cosine warmup scheduling. For post-training, checkpoints are saved every 10,000 steps. For large-scale pre-training, checkpoints are saved every 20,000 steps. We disable progress bars during distributed runs to reduce logging overhead. For packed training, samples are grouped into long sequences under the maximum token budget; samples that exceed the budget are skipped or deferred by the dataloader, while valid shorter samples are packed together to improve token utilization.

At inference time, ERVLA supports reasoning dropout. The model can use explicit CoT, sparsely refreshed CoT, or \texttt{/no\_cot} input depending on the evaluation protocol. Unless otherwise specified, benchmark results are reported under the efficient setting without mandatory test-time CoT decoding.

\section{Detailed Experimental Results}
\label{appendix:detailed_experiments}

\label{appendix: detailed experiments}


\begin{table*}[!t]
\centering
\scriptsize
\setlength{\tabcolsep}{3.8pt}
\renewcommand{\arraystretch}{1.10}
\caption{Ablation of ECoT components on VLABench~\cite{zhang2024vlabenchlargescalebenchmarklanguageconditioned}. The upper part lists the CoT fields used, and the lower part reports performance on different VLABench tracks with and without pretraining on the CoT field data of the Bridge dataset~\cite{walke2024bridgedatav2datasetrobot}.}
\label{tab:ecot_ablation}

\resizebox{\textwidth}{!}{%
\begin{tabular}{ll*{15}{c}}
\toprule
\multicolumn{2}{c}{\textbf{CoT Field}}
& \multicolumn{15}{c}{\textbf{Ablation Setting}} \\
\cmidrule(lr){1-2} \cmidrule(lr){3-17}

\multirow{5}{*}{\textbf{Descriptive}}
& Goal
&   & \checkmark &   &   &   &   &   &   &   & \checkmark &   &   &   &   & \checkmark \\

& Planning
&   &   & \checkmark &   &   &   &   &   &   & \checkmark &   &   &   &   & \checkmark \\

& Subtask
&   &   &   & \checkmark &   &   &   &   &   & \checkmark &   & \checkmark &   & \checkmark & \checkmark \\

& Movement
&   &   &   &   & \checkmark &   &   &   &   & \checkmark &   &   & \checkmark & \checkmark & \checkmark \\

& Reasoning
&   &   &   &   &   & \checkmark &   &   &   & \checkmark &   & \checkmark & \checkmark &   & \checkmark \\

\midrule

\multirow{3}{*}{\textbf{Coordinate}}
& Gripper
&   &   &   &   &   &   & \checkmark &   &   &   & \checkmark &   &   &   & \checkmark \\

& Point traj.
&   &   &   &   &   &   &   & \checkmark &   &   & \checkmark &   &   & \checkmark & \checkmark \\

& Bounding box
&   &   &   &   &   &   &   &   & \checkmark &   & \checkmark &   &   &   & \checkmark \\

\midrule

\multirow{6}{*}{\textbf{w/o Pretraining}}
& In-dist.
& 30.2 & 29.0 & 29.4 & 29.6 & 34.2 & 29.2 & 29.4 & 35.0 & 28.8 & 35.0 & 33.0 & 28.8 & 35.4 & 37.6 & \textbf{38.4} \\

& Category
& 20.2 & 19.0 & 19.4 & 19.6 & 24.4 & 19.2 & 19.6 & 25.0 & 18.8 & 25.0 & 22.8 & 18.8 & 25.4 & 27.6 & \textbf{28.4} \\

& Commonsense
& 7.8 & 6.6 & 7.0 & 7.2 & 12.0 & 6.8 & 7.2 & 12.6 & 6.4 & 12.8 & 10.6 & 6.6 & 13.0 & 15.2 & \textbf{16.0} \\

& Instruction
& 16.4 & 15.2 & 15.6 & 15.8 & 20.4 & 15.4 & 15.6 & 21.2 & 15.0 & 21.4 & 19.0 & 15.2 & 21.6 & 23.8 & \textbf{24.6} \\

& Texture
& 20.4 & 19.2 & 19.6 & 19.8 & 24.4 & 19.4 & 19.6 & 25.2 & 19.0 & 25.2 & 23.0 & 19.0 & 25.6 & 27.8 & \textbf{28.6} \\

& \textbf{Avg.}
& 19.0 & 17.8 & 18.2 & 18.4 & 23.1 & 18.0 & 18.3 & 23.8 & 17.6 & 23.9 & 21.7 & 17.7 & 24.2 & 26.4 & \textbf{27.2} \\

\midrule

\multirow{6}{*}{\textbf{w/ Pretraining}}
& In-dist.
& 35.4 & 34.6 & 34.8 & 34.8 & 37.4 & 34.4 & 29.8 & 36.8 & 29.2 & 38.4 & 33.0 & 33.8 & 38.4 & \textbf{39.0} & 37.8 \\

& Category
& 25.8 & 25.0 & 25.2 & 25.2 & 27.8 & 24.8 & 20.2 & 27.2 & 19.6 & 28.8 & 23.4 & 24.2 & 28.8 & \textbf{29.6} & 28.4 \\

& Commonsense
& 15.2 & 14.4 & 14.6 & 14.4 & 17.2 & 14.2 & 9.6 & 16.6 & 9.2 & 18.4 & 12.8 & 13.6 & 18.2 & \textbf{18.8} & 17.6 \\

& Instruction
& 22.8 & 22.0 & 22.4 & 22.0 & 24.8 & 22.0 & 17.2 & 24.2 & 16.8 & 26.0 & 20.4 & 21.2 & 25.8 & \textbf{26.6} & 25.4 \\

& Texture
& 26.8 & 26.0 & 26.4 & 26.0 & 28.8 & 26.0 & 21.2 & 28.2 & 20.6 & 29.8 & 24.4 & 25.2 & 29.8 & \textbf{30.4} & 29.2 \\

& \textbf{Avg.}
& 25.2 & 24.4 & 24.7 & 24.5 & 27.2 & 24.3 & 19.6 & 26.6 & 19.1 & 28.3 & 22.8 & 23.6 & 28.2 & \textbf{28.9} & 27.7 \\

\midrule

\multirow{6}{*}{\textbf{w/ Pretraining + Dropout}}
& In-dist.
& 35.4 & 34.8 & 35.0 & 34.8 & 37.2 & 34.8 & 34.6 & 38.4 & 34.4 & 38.2 & 36.8 & 34.4 & 38.6 & \textbf{39.8} & 39.4 \\

& Category
& 25.8 & 25.2 & 25.6 & 25.2 & 27.6 & 25.2 & 25.0 & 28.8 & 24.8 & 28.8 & 27.2 & 24.8 & 29.0 & \textbf{30.2} & 29.8 \\

& Commonsense
& 15.2 & 14.6 & 14.8 & 14.6 & 17.2 & 14.6 & 14.4 & 18.2 & 14.2 & 18.2 & 16.6 & 14.2 & 18.4 & \textbf{19.6} & 19.2 \\

& Instruction
& 22.8 & 22.2 & 22.6 & 22.4 & 24.8 & 22.2 & 22.0 & 25.8 & 21.8 & 25.6 & 24.4 & 22.0 & 26.0 & \textbf{27.2} & 26.8 \\

& Texture
& 26.8 & 26.2 & 26.4 & 26.4 & 28.6 & 26.2 & 26.0 & 29.8 & 25.8 & 29.6 & 28.4 & 26.0 & 30.0 & \textbf{31.2} & 30.8 \\

& \textbf{Avg.}
& 25.2 & 24.6 & 24.9 & 24.7 & 27.1 & 24.6 & 24.4 & 28.2 & 24.2 & 28.1 & 26.7 & 24.3 & 28.4 & \textbf{29.6} & 29.2 \\

\bottomrule
\end{tabular}%
}
\end{table*}

\begin{table*}[!t]
\centering
\tiny
\setlength{\tabcolsep}{2.0pt}
\renewcommand{\arraystretch}{1.08}
\caption{Comparison on \textbf{LIBERO} and \textbf{VLABench}. All values are reported in percentage (\%). \textbf{F} denotes Fast, and \textbf{C+F} denotes CoT+Fast. The best result in each column is highlighted in bold.}
\label{tab:libero_liberoplus_combined}

\resizebox{\textwidth}{!}{%
\begin{tabular}{lcccccccccccccccccc}
\toprule
\multirow{3}{*}{\textbf{Model}}
& \multicolumn{8}{c}{\textbf{LIBERO}}
& \multicolumn{10}{c}{\textbf{VLABench}} \\
\cmidrule(lr){2-9} \cmidrule(lr){10-19}
& \multicolumn{2}{c}{\textbf{Spatial}}
& \multicolumn{2}{c}{\textbf{Object}}
& \multicolumn{2}{c}{\textbf{Goal}}
& \multicolumn{2}{c}{\textbf{Long}}
& \multicolumn{2}{c}{\textbf{In-dist.}}
& \multicolumn{2}{c}{\textbf{Category}}
& \multicolumn{2}{c}{\textbf{Common}}
& \multicolumn{2}{c}{\textbf{Instruction}}
& \multicolumn{2}{c}{\textbf{Texture}} \\
\cmidrule(lr){2-3} \cmidrule(lr){4-5} \cmidrule(lr){6-7} \cmidrule(lr){8-9}
\cmidrule(lr){10-11} \cmidrule(lr){12-13} \cmidrule(lr){14-15} \cmidrule(lr){16-17} \cmidrule(lr){18-19}
& F & C+F & F & C+F & F & C+F & F & C+F
& F & C+F & F & C+F & F & C+F & F & C+F & F & C+F \\
\midrule

PaliGemma-2-3B
& 90.2 & 86.4 & 87.6 & 86.0 & 82.4 & 78.4 & 75.8 & 70.8
& 26.8 & 15.4 & 18.6 & 12.2 & 13.8 & 10.2 & 16.2 & 17.0 & 22.2 & 15.2 \\

Florence-2-large
& 87.6 & 86.6 & 90.2 & 86.2 & 78.2 & 78.6 & 72.8 & 70.2
& 32.8 & 15.0 & 13.4 & 11.8 & 19.2 & 10.0 & 12.4 & 16.8 & 15.6 & 15.0 \\

Cosmos-Reason2-2B
& 92.0 & 89.6 & 89.6 & 90.6 & 84.6 & 84.2 & 76.0 & 76.8
& 21.8 & 23.8 & 23.8 & 20.6 & 12.6 & 15.4 & 28.4 & \textbf{33.4} & 18.8 & 22.0 \\

Qwen2.5-VL-3B
& 94.2 & 93.0 & 92.0 & 93.8 & 86.6 & 88.2 & 80.8 & 81.4
& 19.6 & 27.8 & 16.8 & 24.2 & 20.4 & 15.0 & 14.2 & 24.6 & 24.6 & 27.6 \\

Gemma-4-E2B-it
& 90.6 & 92.6 & 93.0 & 93.0 & 85.8 & 87.4 & 78.4 & 80.6
& 34.0 & 37.4 & 21.6 & 28.6 & 16.0 & 20.4 & 21.0 & 30.2 & 17.4 & 29.4 \\

Qwen3-VL-2B
& 95.0 & 95.6 & 94.8 & \textbf{97.4} & 89.6 & 92.0 & 84.8 & 87.2
& 26.8 & 28.8 & 14.6 & 25.2 & 21.2 & 13.9 & 12.6 & 26.2 & 18.2 & 28.8 \\

Qwen2.5-VL-7B
& 91.0 & 95.2 & 94.0 & 95.4 & 87.2 & 90.4 & 80.6 & 84.8
& 28.4 & 37.8 & 15.8 & 28.6 & 14.8 & 20.6 & 18.8 & 30.0 & 14.2 & 30.4 \\

Qwen3-VL-4B
& 95.8 & 96.4 & 96.8 & 96.6 & 91.6 & 91.8 & 86.0 & \textbf{88.6}
& 38.2 & 40.0 & 22.8 & 29.8 & 18.8 & 13.6 & 16.4 & 26.8 & 25.8 & \textbf{31.0} \\

Qwen3-VL-8B
& 90.8 & \textbf{97.2} & 97.0 & 97.2 & 86.4 & \textbf{93.2} & 83.6 & 87.0
& 24.8 & \textbf{40.8} & 16.6 & \textbf{32.4} & 15.2 & \textbf{22.6} & 13.6 & 25.2 & 20.0 & 29.2 \\

\bottomrule
\end{tabular}%
}
\end{table*}

\subsection{Simulation Results}
\label{appendix:detailed_simulation_results}

\begin{figure}[t]
    \centering
    \includegraphics[width=1.0\textwidth]{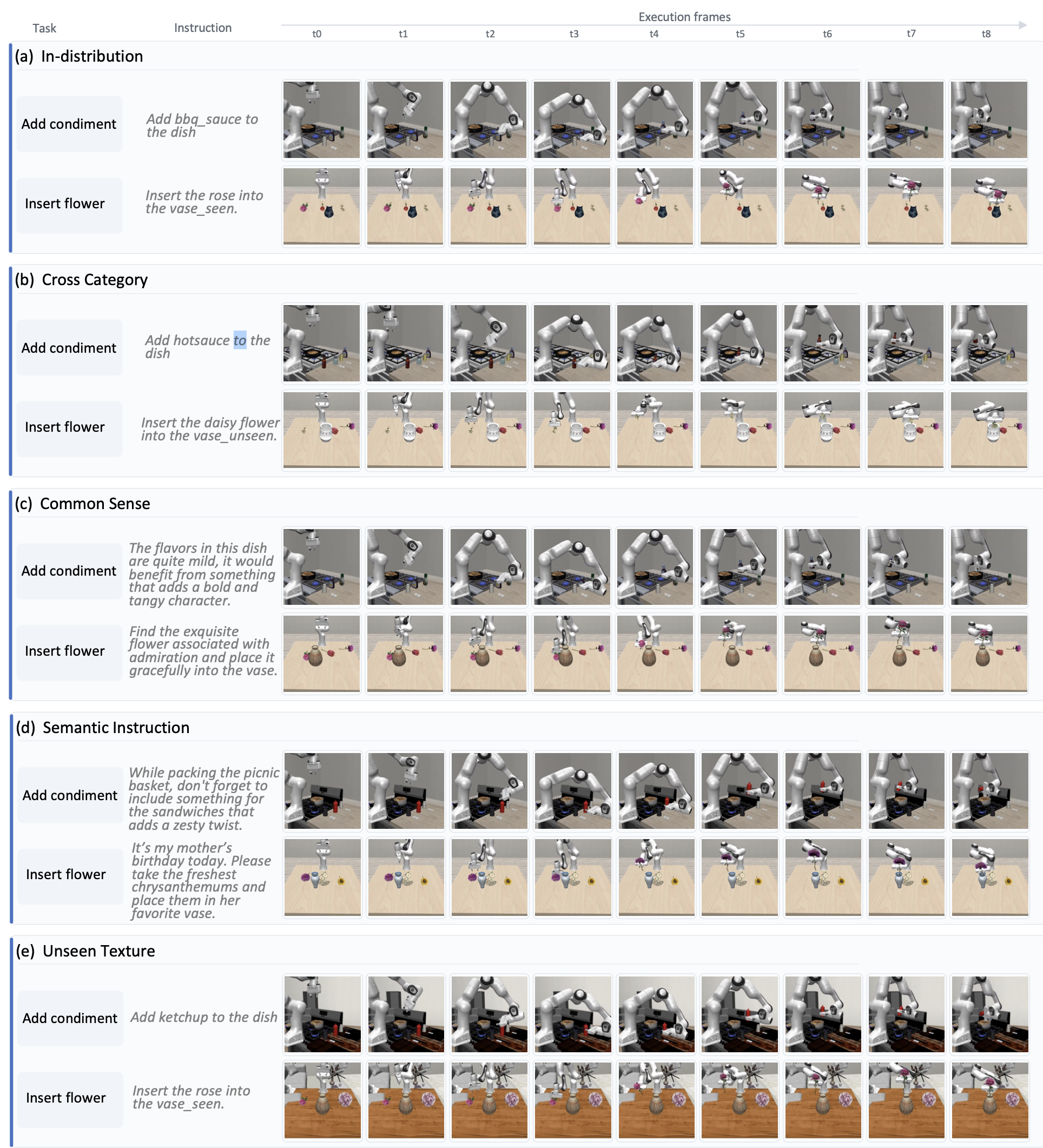}
\caption{
\textbf{Visualization of VLABench evaluation results.}
}
    \label{fig:vlabench_sim}
\end{figure}

\begin{figure}[t]
    \centering
    \includegraphics[width=1.0\textwidth]{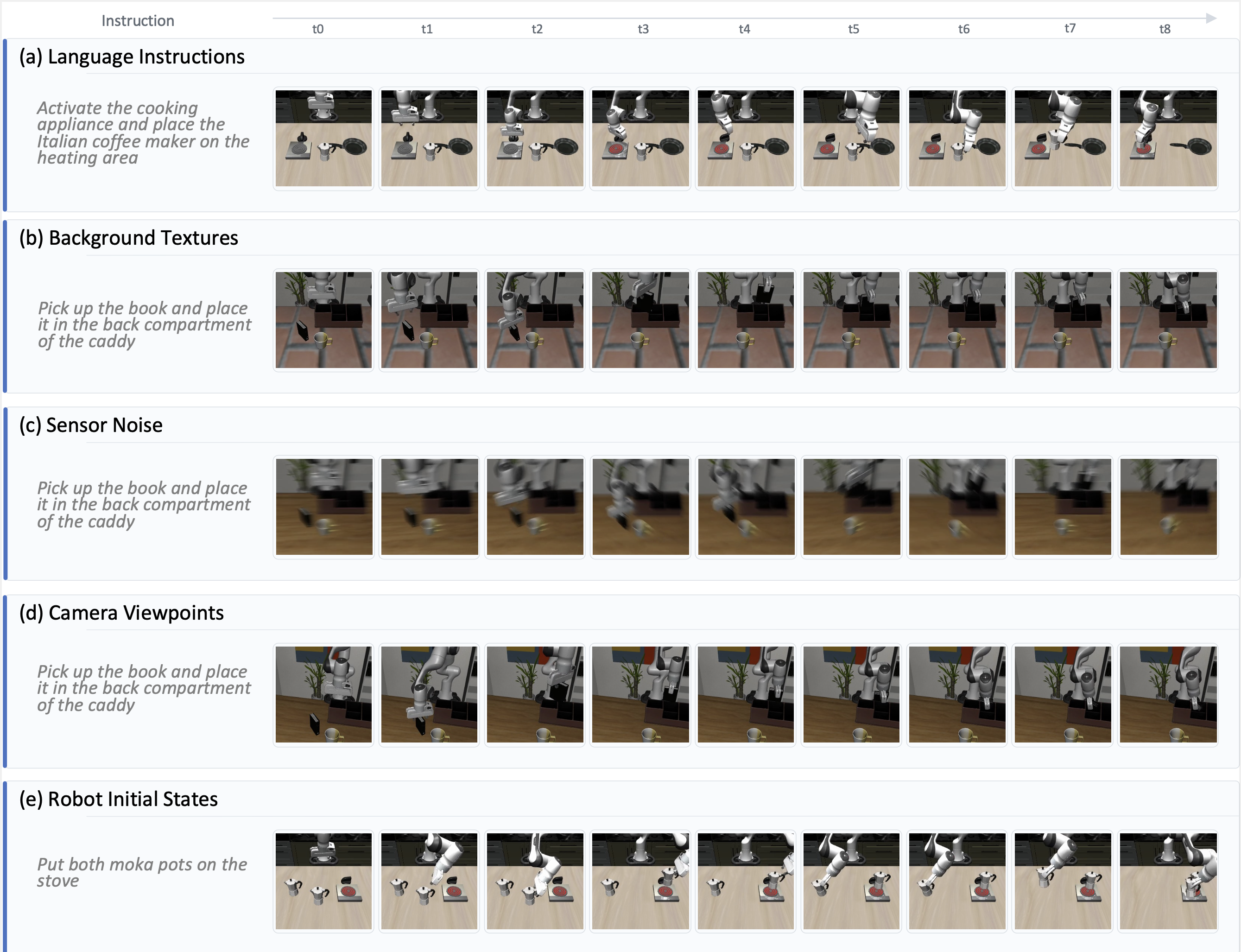}
\caption{
\textbf{Visualization of LIBERO-PLUS evaluation results.}
}
    \label{fig:liberoplus_sim}
\end{figure}

This section provides the full simulation results corresponding to the controlled studies and benchmark comparisons in Sec.~\ref{sec:explore} and Sec.~\ref{sec:study}. We report three complementary analyses. First, Table~\ref{tab:ecot_ablation} gives the full field-level ECoT ablation on VLABench~\cite{zhang2024vlabenchlargescalebenchmarklanguageconditioned}. Second, Table~\ref{tab:libero_liberoplus_combined} reports the VLM-to-VLA transfer results across LIBERO~\cite{liu2023liberobenchmarkingknowledgetransfer} and VLABench under the same FAST action interface. Third, Table~\ref{tab:real_world} reports the detailed real-world breakdown used in Sec.~\ref{sec:real-world}; although it is not a simulation benchmark, we include it here for completeness so that all detailed quantitative results are collected in one appendix section. We visualize representative simulation rollouts in Figures~\ref{fig:vlabench_sim} and~\ref{fig:liberoplus_sim}.

\paragraph{Field-level ECoT ablations on VLABench.}
Table~\ref{tab:ecot_ablation} expands the analysis in Sec.~\ref{sec:explore} by reporting every CoT-field configuration across five VLABench tracks. All variants use the same Qwen3-VL-4B backbone and the same autoregressive CoT+FAST action-token interface, so the comparison isolates the effect of the reasoning fields. The upper block specifies which ECoT fields are enabled, while the lower blocks report results under three regimes: direct post-training without Bridge pre-training, Bridge-based ECoT pre-training, and Bridge-based ECoT pre-training with reasoning dropout. The results show a clear pattern: high-level descriptive fields such as goal, planning, subtask, and reasoning are not sufficient when used alone, whereas action-centric fields such as movement description and point trajectory provide stronger gains. However, dense coordinate fields such as gripper position and bounding boxes become harmful under pre-training when noisy labels are repeatedly imposed, revealing CoT contamination. Reasoning dropout mitigates this degradation by preventing the model from over-relying on explicit and potentially noisy CoT fields.

\paragraph{VLM-to-VLA transfer under a fixed action interface.}
Table~\ref{tab:libero_liberoplus_combined} provides the detailed results for the VLM-to-VLA transfer study. Each backbone is trained with the same StarVLA protocol and FAST action-token interface, and each method is evaluated with and without full ECoT supervision. The LIBERO results show that many VLMs already achieve strong in-distribution performance, but ECoT improves the stronger Qwen-series backbones more consistently, especially on object, goal, and long-horizon task suites. The VLABench results are more discriminative: without CoT, stronger VLMs do not always transfer reliably to harder semantic and out-of-distribution tracks; with CoT+FAST, the Qwen3-VL series obtains clearer gains, especially on in-distribution, cross-category, commonsense, instruction, and texture generalization tracks. This supports our conclusion that embodied CoT acts as a transfer interface: it helps convert pretrained semantic priors into action-relevant representations, rather than merely adding extra text supervision.

\paragraph{Connection to main benchmark results.}
The detailed tables also clarify why ERVLA is designed beyond autoregressive CoT+FAST. Table~\ref{tab:ecot_ablation} shows that explicit CoT is useful only when the right fields are selected and noisy grounding is regularized. Table~\ref{tab:libero_liberoplus_combined} shows that ECoT improves VLM-to-VLA transfer, but still remains constrained by autoregressive action decoding. These observations motivate the ERVLA interface evaluated in Tables~\ref{tab:libero_plus_comparison} and~\ref{tab:vlabench_comparison}: CoT is used as representation-shaping supervision, while continuous action generation is handled through the choice branch, knowledge truncation, and DiT-based flow modeling. Together, the detailed simulation results support the main claim that embodied CoT becomes scalable when it is internalized into action-aware representations rather than treated as a mandatory action prefix.

\begin{table}[!t]
\centering
\small
\setlength{\tabcolsep}{4.5pt}
\renewcommand{\arraystretch}{1.08}
\caption{\textbf{Real-world task suite.} We evaluate 20 tasks across four difficulty tiers. Each task is tested with five trials.}
\label{tab:real_world_tasks}
\resizebox{\linewidth}{!}{
\begin{tabular}{cll}
\toprule
\textbf{Tier} & \textbf{Task family} & \textbf{Instruction} \\
\midrule

\multirow{5}{*}{Basic}
& Drawer placing & Put the blue toy car into the drawer. \\
& Drawer placing & Put the red toy car into the drawer. \\
& Drawer placing & Put the yellow toy car into the drawer. \\
& Table clearing & Put the empty can into the basket. \\
& Table clearing & Put the bottle into the basket. \\

\midrule

\multirow{5}{*}{Distractors}
& Drawer placing & Put the toy car into the drawer. (with fruit models) \\
& Drawer placing & Put the blue toy car into the drawer. (with other toy cars) \\
& Drawer placing & Put the toy car into the drawer. (with cans and bottles) \\
& Table clearing & Put the can into the basket. (with the fruits and other bottles) \\
& Table clearing & Clear the bottle from the table. (with toy cars) \\

\midrule

\multirow{5}{*}{Semantic}
& Drawer placing & Put away the object that is not a fruit. \\
& Drawer placing & Put the toy into the lower drawer. \\
& Drawer placing & Put the toy into the drawer above the lower one. \\
& Table clearing & Remove the item that should be thrown away. \\
& Table clearing & This table is too messy. \\

\midrule

\multirow{5}{*}{Long-horizon}
& Drawer placing & Put the non-toy items into the upper drawer, and put the toys into the second drawer. \\
& Drawer placing & Put all toy cars into the upper drawer while leaving the fruits on the table. \\
& Drawer placing & Put all fruits into the lower drawer. \\
& Table clearing & Clear all waste items from the table and place them into the basket. \\
& Clearing and Placing  & Put the non-toy items into the upper drawer, and put the toys into the second drawer. \\

\bottomrule
\end{tabular}}
\vspace{-2mm}
\newpage
\end{table}

\begin{figure}[t]
    \centering
    \includegraphics[width=1.0\textwidth]{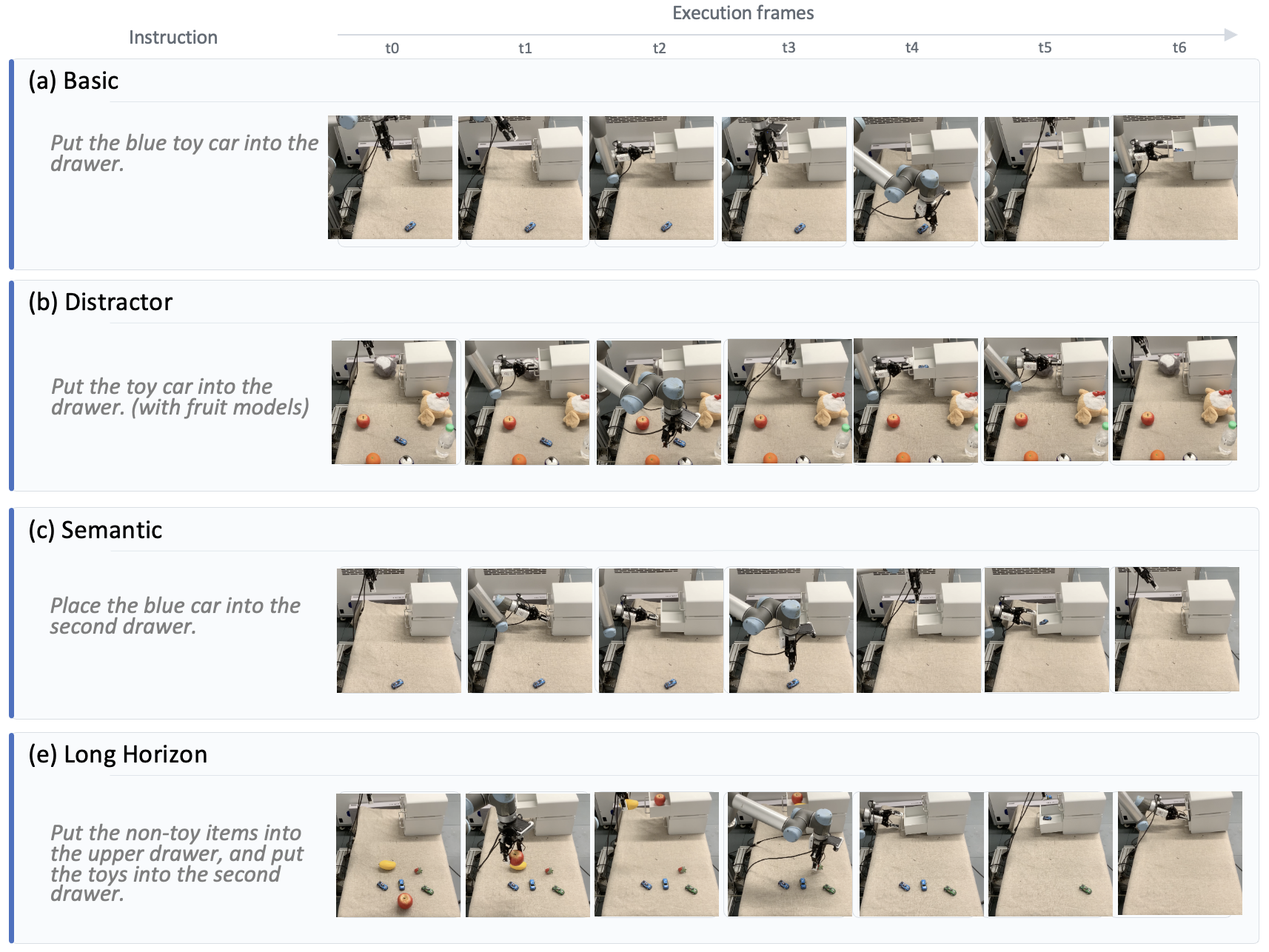}
\caption{
\textbf{Visualization of real-world evaluation results.}
}
    \label{fig:real_case}
\end{figure}

\subsection{Real-world Results}
\label{appendix:real_world_results}

We provide the complete real-world task suite and evaluation results in Tables~\ref{tab:real_world_tasks} and~\ref{tab:real_world}. 
The robot platform and representative rollouts are described in Sec.~\ref{sec:real-world}; here we additionally report the full task instructions, evaluation protocol, training setup, and quantitative breakdown across difficulty tiers. 
As shown in Table~\ref{tab:real_world_tasks}, each method is evaluated on 20 tasks spanning \textit{Basic}, \textit{Distractors}, \textit{Semantic}, and \textit{Long-horizon} settings, with five trials per task, resulting in 100 real-world rollouts per method. 
We report two metrics in Table~\ref{tab:real_world}: success rate (SR), which measures whether the full instruction is completed, and progress score (PS), which assigns partial credit for correctly completed subtasks such as opening the drawer, selecting the correct object, placing an object into the target container, or completing one stage of a multi-object instruction. 
This two-metric protocol is important because many failures in real-world manipulation are not binary: a policy may correctly identify the target object but fail to place it, complete the first object in a multi-object instruction but lose track of the remaining objects, or execute the correct motion primitive but choose an incorrect container.

We collect approximately 10 hours of demonstrations covering object pick-and-place with toy cars, fruit models, cans, bottles, and drawer opening and closing. 
The real-world demonstrations are annotated with the same embodied CoT schema used in the main simulation experiments, including task understanding, object grounding, planning, subtask reasoning, movement description, gripper pose, and future waypoint trajectory when available. 
This keeps the real-world adaptation setting consistent with our pre-training and benchmark post-training setup, so the comparison reflects whether the learned reasoning-action interface transfers to physical execution rather than whether a new real-world prompt format is introduced. 
All models are fine-tuned on the same real-world demonstrations using the same optimization schedule: 8 NVIDIA A100 GPUs, per-GPU batch size 16, and 20,000 training steps. 
To reduce execution jitter and make continuous-action rollouts comparable, we apply the same temporal action ensemble strategy~\cite{li2024cogactfoundationalvisionlanguageactionmodel} to all continuous-action methods.

The four difficulty tiers separate different sources of real-world generalization error. 
\textit{Basic} evaluates clean pick-and-place and drawer interaction with explicit object names. 
\textit{Distractors} adds irrelevant or visually similar objects to test target selectivity under clutter. 
\textit{Semantic} replaces direct object names with category, function, or spatial-reference descriptions, requiring the policy to infer the intended object or container. 
\textit{Long-horizon} combines semantic grounding with multi-object, multi-step execution, testing whether the policy can preserve progress over an extended task.

Table~\ref{tab:real_world} shows that the main performance gap appears beyond clean manipulation. 
On the \textit{Basic} tier, $\pi_{0.5}$ and ERVLA perform similarly, reaching 97/98 and 96/97 SR/PS, respectively, suggesting that strong continuous-action policies already handle simple low-level manipulation well. 
However, under \textit{Distractors}, \textit{Semantic}, and \textit{Long-horizon} settings, the policy must resolve ambiguity, ignore irrelevant objects, and maintain task intent over time, making reasoning-aware representations more important.

ERVLA shows the clearest gains on semantic and long-horizon tasks (see Figure.~\ref{fig:real_case}). 
On the \textit{Semantic} tier, it improves over $\pi_{0.5}$ from 31 to 42 SR and from 45 to 58 PS, indicating better conversion of semantic intent into action-relevant representations. 
On the \textit{Long-horizon} tier, ERVLA improves PS from 38 to 55, showing that it more often completes meaningful intermediate stages even when the full instruction is not perfectly solved. 
This supports our design: embodied CoT pre-training provides subtask and movement-level structure, while reasoning dropout encourages these signals to be internalized rather than emitted as fragile test-time text.

Overall, the real-world results show that explicit reasoning alone is insufficient: ECoT suffers from autoregressive latency and error propagation, while visual or latent reasoning in WorldVLA and UniVLA can obscure fine-grained semantic grounding. 
ERVLA instead uses embodied CoT as training supervision, reasoning dropout to avoid dependence on explicit inference-time traces, and cache-conditioned continuous action generation to preserve both semantic grounding and smooth control.

\section{Limitations and Future Work}

Our work shows that embodied CoT is most useful when it is absorbed into action-relevant representations rather than exposed as a mandatory verbal trace. This also reveals a central limitation: the value of reasoning pre-training is bounded by the quality of the reasoning substrate. In our pipeline, goals, plans, movement descriptions, object boxes, gripper locations, and point trajectories provide a structured bridge from semantics to control, but they are not equally reliable. Linguistic fields are usually robust but sometimes underspecified; dense grounding fields are more directly actionable but more vulnerable to detector errors, calibration bias, and occlusion. Reasoning dropout and knowledge truncation reduce the damage of imperfect supervision, yet they do not remove the need for better annotation quality. A natural next step is to make uncertainty part of the annotation itself, so that the model learns not only what to reason about, but also when each reasoning signal should be trusted.

Our study also treats reasoning primarily as an offline pre-training signal. This makes inference efficient and avoids mandatory test-time CoT decoding, but leaves open how much reasoning should be dynamically refreshed during execution. Long-horizon manipulation may require memory of completed subtasks, recovery from failed attempts, or clarification when the scene does not match the instruction. Our results suggest that the answer is unlikely to be simply ``more explicit reasoning.'' Future systems should decide when to verbalize, when to internalize, and when to act directly. In this sense, our work points toward a practical form of embodied reasoning: not a robot that talks through every motion, but one whose actions carry the structure of reasoning even when no reasoning text is produced.



\end{document}